\documentclass[journal]{IEEEtran}
\usepackage{amsmath,amsfonts}
\usepackage{algorithmicx}
\usepackage{algorithm}
\usepackage{algpseudocode}
\usepackage{array}
\usepackage[caption=false,font=normalsize,labelfont=sf,textfont=sf]{subfig}
\usepackage{textcomp}
\usepackage{stfloats}
\usepackage{url}
\usepackage{verbatim}
\usepackage{graphicx}
\usepackage{cite}
\usepackage{booktabs}
\usepackage{amssymb}
\usepackage[pagebackref,breaklinks=true,colorlinks,citecolor=blue,urlcolor=blue,linkcolor=blue]{hyperref}
\usepackage[section]{placeins}
\usepackage{bm}
\usepackage{balance}
\usepackage[capitalize]{cleveref}
\usepackage{multirow}
\usepackage{makecell}
\usepackage{pifont}
\usepackage{enumitem}
\usepackage{caption} %% 使得TABLE I 之后不换行
\usepackage{orcidlink}
\usepackage{pifont}  %% \cmark需要
\usepackage{xspace} %% \cmark需要，智能添加空格
\usepackage{xcolor}
\newcommand{\cmark}{\ding{51}\xspace}%
\newcommand{\xmarkg}{\textcolor{lightgray}{\ding{55}}\xspace}%

\usepackage{color}
\usepackage{colortbl}
\newif \ifshowchanges
\showchangesfalse   % final version, black

\ifshowchanges
  \newcommand{\re}[1]{\textcolor{blue}{#1}}  % 蓝色字体版本
\else
  \newcommand{\re}[1]{#1}                  % 正常黑色版本
\fi

\begin{document}

\title{PAGE: \textbf{P}arts-\textbf{A}ware \textbf{G}uidanc\textbf{e} for Co-Speech Gesture Portrait Video Generation}

\author{
Yabiao Wang\orcidlink{0000-0002-6592-8411}, Jiafu Wu, Yicheng Xu, Teng Hu, Shengzhe Zhou, Yanxiao Sun,\\
Jiangning Zhang, Ying Tai, Yong Liu, and Ming-Hsuan Yang,~\IEEEmembership{Fellow,~IEEE}

\thanks{Yabiao Wang, Yong Liu are with the Institute of Cyber-Systems and Control, Zhejiang University, Hangzhou 310027, China (e-mail: yabiaowang@zju.edu.cn; yongliu@iipc.zju.edu.cn).}
\thanks{Yabiao Wang and Jiafu Wu contributed equally to this work. Corresponding author: Yong Liu.}
\thanks{Digital Object Identifier 10.1109/TCSVT.2026.3738661} 

}

\markboth{IEEE Transactions on Circuits and Systems for Video Technology}{Wang \MakeLowercase{\textit{et al.}}: PAGE: Parts-Aware Guidance for Co-Speech Gesture Portrait Video Generation}

\maketitle

\begin{abstract}
Co-speech gesture video generation aims to synthesize realistic, audio-driven portrait videos from a single reference image and a speech signal. Despite advances in visual quality, diffusion-based approaches still suffer from degraded local details and imperfect audio–visual alignment.
We present PAGE (\textit{Parts-Aware GuidancE}), a general, backbone-agnostic optimization framework for diffusion-based generation that enhances local spatial fidelity and global audio–visual consistency at both the \textit{training-objective} and \textit{denoising-trajectory} levels. During training, we propose \textbf{Parts-Specific Reweighting (PSR)}, which adaptively increases supervision in semantically important regions such as lips, eyes, and hands, improving visual fidelity in perceptually critical areas. During inference, we introduce \textbf{Consistency Classifier Guidance (CCG)}, an audio-conditioned latent classifier that identifies and corrects audio–visual inconsistencies in the generated video. We further design \textbf{Differential Guidance (DG)}, an efficient inference strategy that refines facial and body motion separately while avoiding cross-region interference.
To enable large-scale evaluation, we construct the \textbf{Chinese Announcer Speech (CAS)} dataset—the first comprehensive Chinese benchmark for co-speech gesture portrait video generation. Compared with the widely used PATS dataset, CAS includes substantially more identities, video clips, and scene diversity. Extensive experiments on both PATS and CAS demonstrate that PAGE achieves state-of-the-art performance in perceptual quality, gesture diversity, and audio–visual alignment. Project page: \url{https://aigc-explorer.github.io/PAGE/}.
\end{abstract}


\section{Introduction}

\IEEEPARstart{A}{udio}-driven portrait animation aims to synthesize realistic and temporally coherent videos of a speaking person from a single reference image and speech audio. This technology has broad applications in film production, digital humans, gaming, virtual presenters, and online education.

\begin{figure}[t]
    \centering
    \includegraphics[width=\linewidth]{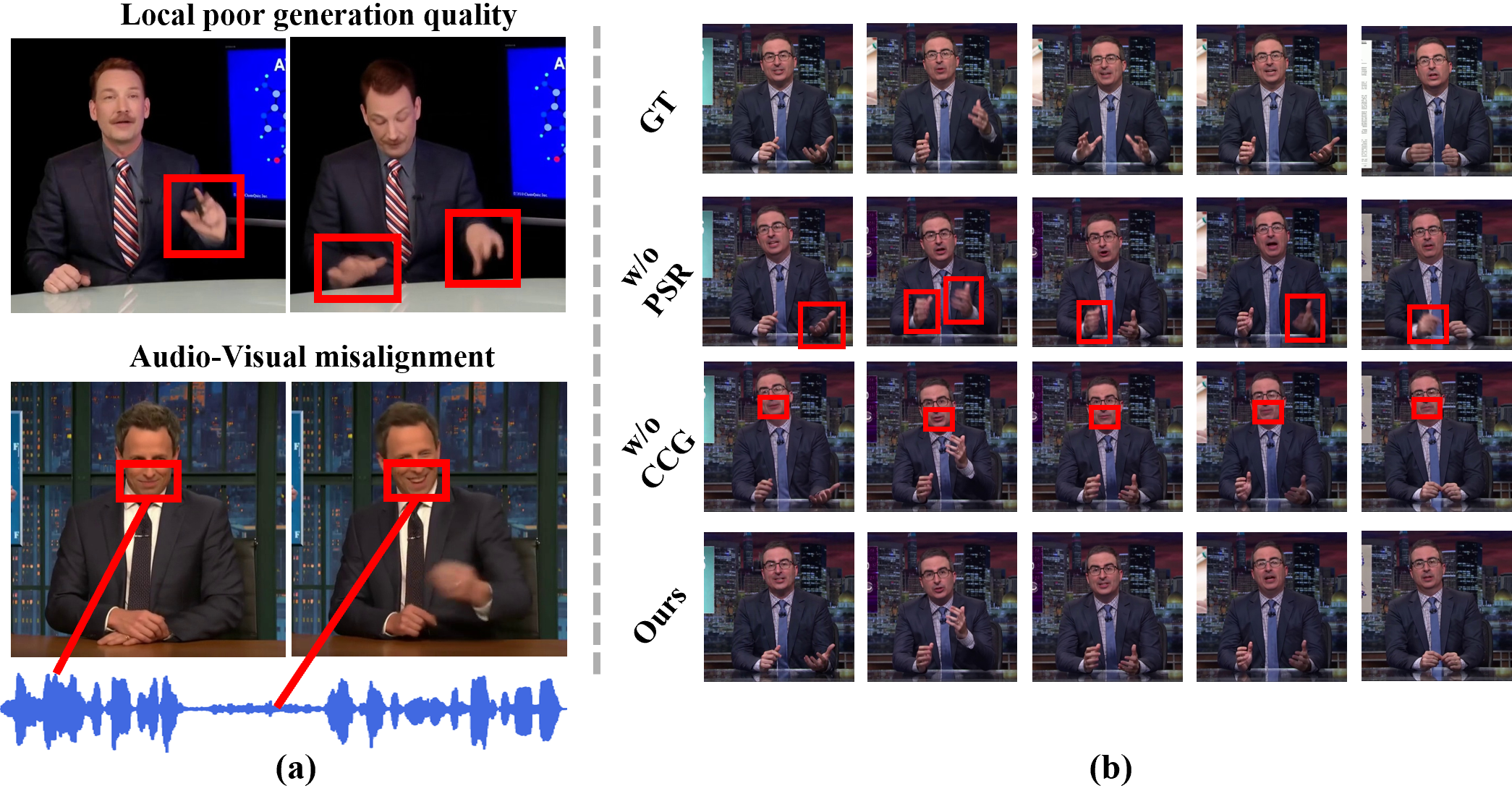}
    \caption{ 
\textbf{\textit{(a)}} \textbf{Limitations of existing methods} include degraded local fidelity (e.g., hands, facial regions) and audio–visual misalignment (e.g., motion that does not match speech).
\textbf{\textit{(b)}} \textbf{Qualitative ablation.} Parts-Specific Reweighting (PSR) improves local visual quality in critical regions, while Consistency Classifier Guidance (CCG) refines the generated sequence at inference time to enhance audio–visual alignment. The full PAGE framework combines both, producing high-quality and temporally consistent co-speech gesture videos.
    }
\label{fig:teaser} 
\end{figure}

Existing methods for this task can be broadly grouped into three categories: VAE-based~\cite{liu2022audio,qian2021speech}, GAN-based~\cite{3DGestureGAN,ginosar2019learning,guo2024liveportrait}, and diffusion-based approaches~\cite{EMO2024,ji2025sonic,xu2024vasa,wei2024aniportrait,he2024S2G,meng2025echomimicv2,hu2025hunyuancustom}.  
VAE-based pipelines first predict intermediate gesture or pose representations and then render them into images~\cite{balakrishnan2018gesture2image,chan2019everybody_dace}. These multi-stage frameworks often yield limited realism and require complex supervision.  
GAN-based methods directly synthesize images with adversarial training, producing sharper results but suffering from instability and limited diversity.  

Recently, diffusion-based models have achieved substantial improvements in portrait animation. Systems such as EMO~\cite{EMO2024}, Sonic~\cite{ji2025sonic}, and VASA-1~\cite{xu2024vasa} generate high-fidelity talking-head videos directly from audio. However, most focus on the face or head, neglecting body and hand gestures that are essential for natural co-speech behavior. For upper-body animation, S2G~\cite{he2024S2G} employs thin-plate spline transformations to guide motion synthesis. EchoMimic~\cite{meng2025echomimicv2} and Hallo3~\cite{cui2025hallo3} leverage large-scale diffusion backbones. Nevertheless, as illustrated in Fig.~\ref{fig:teaser}, these models still face two major challenges:

\begin{itemize}
    \item \textbf{Local degradation in critical regions.} Lips, eyes, and hands are perceptually salient but prone to artifacts such as blurring, distortion, or misalignment. 
    \item \textbf{Audio–visual misalignment.} The generated motion often fails to follow the rhythm, energy, or semantics of the speech signal, especially over long sequences.
\end{itemize}

To address these issues, we propose \textbf{PAGE} (\textit{Parts-Aware Guidance}), a unified diffusion-based framework that improves spatial detail and temporal alignment. PAGE adopts a divide-and-conquer strategy to separately enhance facial and body gesture synthesis while maintaining coherent overall motion.

First, we introduce \textbf{Parts-Specific Reweighting (PSR)} (Sec.~\ref{sec:psr}), which adaptively increases loss weights on semantically important regions—lips, eyes, and hands—using pose confidence and segmentation masks. This encourages the generator to allocate higher representational capacity to visually critical areas.  
Second, during inference, we propose \textbf{Consistency Classifier Guidance (CCG)} (Sec.~\ref{sec:CCG}), an audio-conditioned latent classifier that predicts whether the generated video aligns with the speech. Its gradients are used to refine generation, enhancing synchrony without retraining.  
Finally, we develop \textbf{Differential Guidance (DG)} (Sec.~\ref{sec:DG}), an efficient inference strategy that applies face and non-face area guidance independently, preventing interference and improving consistency at negligible additional cost.

To enable large-scale evaluation of co-speech gesture video generation, we construct the \textbf{Chinese Announcer Speech (CAS)} dataset (Sec.~\ref{sec:dataset}). CAS is the first large-scale Chinese dataset for this task, containing over 19,000 clips and 800 speakers with diverse scenarios, camera views, and gestures.

\re{PAGE is designed as a general optimization framework for diffusion-based co-speech gesture portrait video generation, operating at both the \textbf{training-objective} and \textbf{denoising-trajectory} levels. Its design is orthogonal to specific diffusion backbone architectures and does not rely on any particular diffusion model.
Our main contributions are summarized as follows:}
\begin{itemize}
    \item We propose \textbf{PAGE}, a unified diffusion-based framework for co-speech gesture portrait video generation that enhances both spatial fidelity and audio–visual alignment.
    \item We introduce \textbf{Parts-Specific Reweighting (PSR)} to emphasize semantically important regions, improving local detail and perceptual realism.
    \item We design \textbf{Consistency Classifier Guidance (CCG)} and \textbf{Differential Guidance (DG)} to refine alignment during inference, which is effective and efficient.
    \item We construct the \textbf{Chinese Announcer Speech (CAS)} dataset, a large-scale and diverse benchmark for Chinese co-speech gesture generation.
    \item Extensive experiments on PATS and CAS demonstrate that PAGE achieves state-of-the-art results in quality, diversity, and synchrony.
\end{itemize}

\section{Related Work}
\subsection{Audio-Driven Portrait Animation}
Audio-driven portrait animation methods can be divided into
two categories: talking face/head generation and co-speech
gesture generation (body portrait animation).

\noindent\textbf{Talking-face and head animation.}
This research focuses on generating facial motion synchronized with speech audio. Early methods relied on 3D morphable models or landmarks, while recent works use neural representations for realistic animation.  
% \wyb{Eskimez \textit{et al.}~\cite{eskimez2021speech} generated emotion-aware talking faces using audio and emotion labels.} \wyb{Ye \textit{et al.}~\cite{ye2022audio} introduced dynamic convolution kernels for real-time high-quality generation.}  
% \wyb{Anyoneet~\cite{wang2022anyonenet} synthesizes synchronized speech and video from text and a single reference image.}
SLIGO~\cite{SLIGO2024Sheng} introduces a deep state-space model optimized via DVAE with CNF prior to capture stochastic audio-driven facial dynamics. TH-Mamba~\cite{THMamba2025xu} proposes Mamba-based spatiotemporal fusion with intra/inter-frame modules and composite constraints, aligning audio-image to reduce distortions.
OSM-Net~\cite{liu2024osm} proposes a one-to-many one-shot talking head model that maps audio into a learned clip-level motion space to sample natural, diverse head motions. Liu \textit{et al.}~\cite{ASE2024liu} introduce an audio-semantic enhanced, pose-driven talking head generator leveraging multimodal 3DMM prediction. 
% Song \textit{et al.}~\cite{song2022audio} propose an audio-driven UGC dubbing method that rapidly adapts to unseen speakers via cross-modal style-content fusion.
VASA-1~\cite{xu2024vasa} uses 3D-aware latent representations to disentangle identity and motion for consistent talking-head synthesis.  
% ConsistentAvatar~\cite{yang2024consistentavatar} and PersonaTalk~\cite{zhang2024personatalk} improve temporal consistency and stylistic control via temporal diffusion and audio cross-attention.  
Recent works such as TalkinNeRF~\cite{chatziagapi2024talkinnerf} and EMO~\cite{EMO2024} integrate 3D scene representations and end-to-end diffusion to enhance realism and identity preservation.  
However, these methods primarily target the face, lacking expressive upper-body gestures crucial for natural communication.

\noindent\textbf{Co-speech gesture generation.}
Co-speech gesture synthesis extends beyond facial motion to include head, torso, and hand gestures aligned with speech.  
ANGIE~\cite{liu2022audio} and SDT~\cite{qian2021speech} model audio-to-pose mapping via learned motion primitives or template vectors.  
S2G~\cite{he2024S2G} and DiffTED~\cite{hogue2024diffted} incorporate temporal dynamics using spline-based motion representations.  
Make-Your-Anchor~\cite{huang2024make} adds controllable 3D mesh constraints, while EchoMimicV2~\cite{meng2025echomimicv2} and Hallo3~\cite{cui2025hallo3} use diffusion-based audio–pose harmonization and partial attention mechanisms. MMGT~\cite{mmgt2025wang} proposes a two-stage audio-driven framework generating videos and using motion-masked audio attention in diffusion to capture large, synchronized co-speech gestures. 
% Despite progress, these models still exhibit limited fidelity in hands and lips and inconsistent synchrony between gesture and prosody. PAGE directly addresses these limitations with explicit part-aware training and inference guidance.

\subsection{Diffusion-Based Video Generation}
Diffusion models have demonstrated strong performance in image synthesis\cite{zhang2023adding,mokady2023null}, gesture generation~\cite{wang2025timotion,wang2026marrs}, and video synthesis~\cite{kong2025hunyuanvideo,wan2025wan}. 
VDM~\cite{ho2022video} and SVD~\cite{blattmann2023SVD} extend diffusion to spatiotemporal domains, while AnimateDiff~\cite{guo2024animatediff} introduces motion modules to adapt text-to-image models for animation.  
Human-centric systems such as Animate Anyone~\cite{hu2024animate} and Champ~\cite{zhu2025champ} enable pose-controllable video synthesis.  
Large-scale diffusion transformers (DiT) such as HunyuanVideo~\cite{kong2025hunyuanvideo} and Wan~\cite{wan2025wan} further enhance temporal coherence through global spatiotemporal attention and large-scale pretraining.  
PAGE builds on these advances, integrating region-aware training and audio-conditioned inference for co-speech gesture synthesis.

\section{Method}

\begin{figure*}[tbp]
    \centering
    \includegraphics[width=\linewidth]{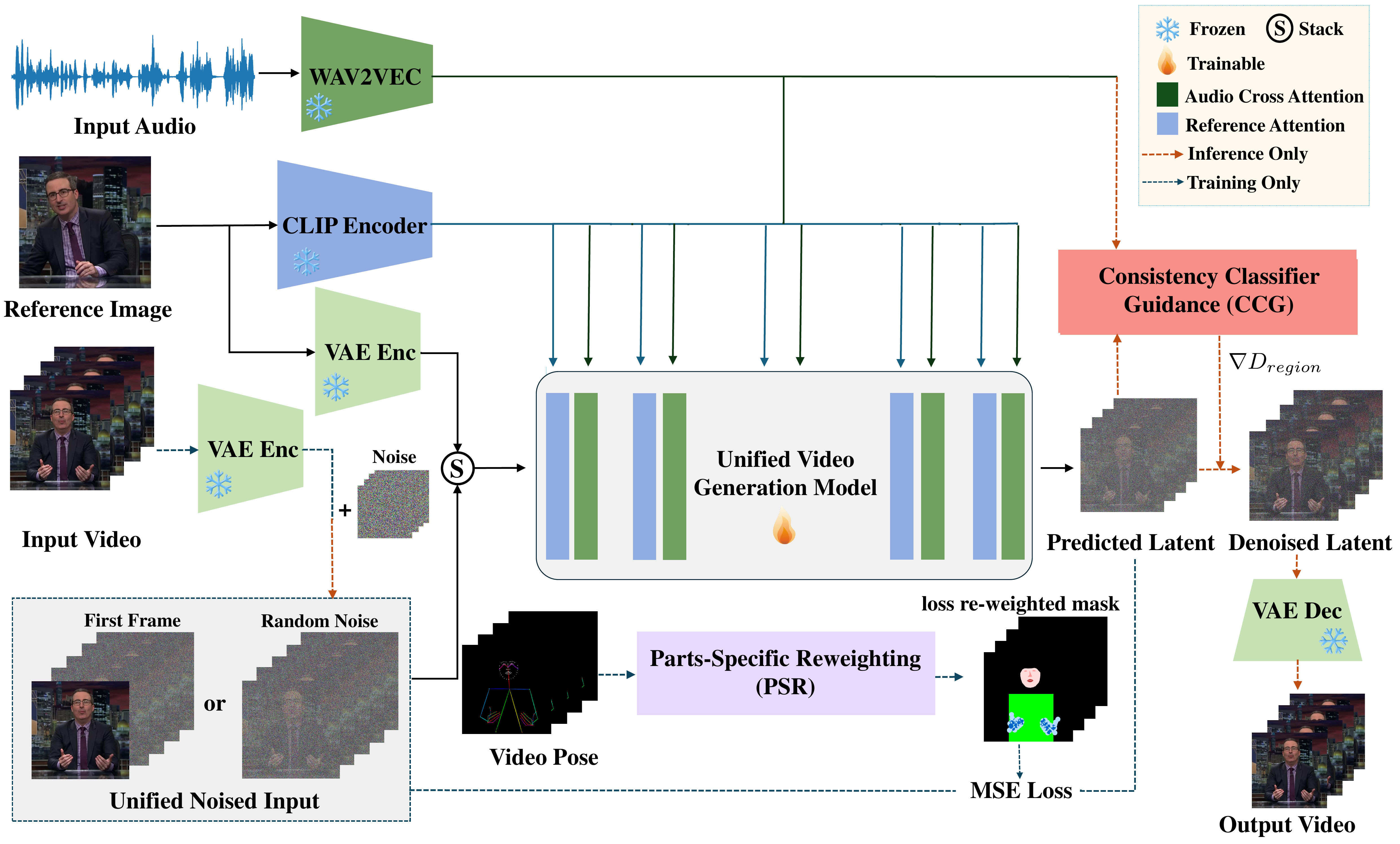}
    \caption{\textbf{Overview of the proposed PAGE} with three components:
    \textbf{(a) Backbone}: a video generation model {$\boldsymbol{G}$} (e.g., Wan2.1). Video frames are encoded with a VAE; reference-image latents are extracted using CLIP and the VAE. These features, together with noise from a conditional first frame or a noised video clip, are concatenated along the channel dimension. Audio embeddings from a pretrained audio encoder are injected into {\boldsymbol{$G$}} via cross-attention.
    \textbf{(b) Parts-Specific Reweighting (PSR)}: a dynamic, region-aware loss weighting scheme constructed from pose-keypoint confidences, which upweights supervision for critical regions during training.
   \textbf{(c) Consistency Classifier Guidance (CCG)}: at inference, consistency gradients derived from joint audio–video features guide denoising to improve temporal audio–visual coherence. The VAE decoder then maps denoised latents back to pixel space.}
    \label{pipeline}
\end{figure*}

We propose \textbf{PAGE}, illustrated in Fig.~\ref{pipeline}, a co-speech gesture portrait video generation framework consisting of two core components: Parts-Specific Reweighting (PSR), which is applied during training, and Consistency Classifier Guidance (CCG), which is applied during inference. PSR enforces spatially adaptive supervision so that the model focuses on semantically important and challenging regions such as lips, eyes, and hands. During inference, CCG uses two consistency classifiers and Differential Guidance (DG) to deliver decoupled, region-aware gradients. This substantially enhances the visual fidelity of dynamic regions and improves temporal alignment between motion and speech.

\subsection{Preliminaries}

\noindent\textbf{Diffusion models.}
Diffusion Transformers (DiT)~\cite{peebles2023dit} extend denoising diffusion models by replacing the convolutional U-Net denoiser with a Transformer-based backbone. Given a clean latent $z$, Gaussian noise $\epsilon$, and a timestep $t$, a noised latent is constructed as
\begin{equation}
    z_t = \sqrt{\alpha_t} z + \sqrt{1-\alpha_t}\,\epsilon,
\end{equation}
where $\alpha_t$ controls the signal-to-noise ratio at step $t$. The denoiser $\epsilon_\theta$ is trained to predict the injected noise via
\begin{equation}
    \mathcal{L} = 
    \mathbb{E}_{t, z_t, c, \epsilon \sim \mathcal{N}(0, 1)}
    \left[
        \| \epsilon_\theta(z_t, t, c) - \epsilon \|_2^2
    \right],
\label{eq:ori_flow}
\end{equation}
where $c$ denotes conditioning information (e.g., reference frame, audio). The Transformer-based denoiser uses full spatiotemporal attention, which is beneficial for long-range temporal modeling in videos.

In our framework, we adopt Wan2.1 as the video generator $\boldsymbol{G}$. Wan2.1 is a diffusion Transformer built on the flow matching\cite{lipman2023flow} paradigm, incorporating full attention for temporal modeling. This architecture enables high-fidelity, temporally consistent video synthesis over long durations.

\noindent\textbf{Audio embedding.}
Given an input audio sequence of length $T$, we extract speech representations using wav2vec~\cite{baevski2020wav2vec}. The corresponding video clip of length $T$ is encoded by a pretrained 3D VAE with a temporal compression ratio of 4, producing $(T+3)/4$ latent frames where the factor of 4 is the time compression ratio of the 3D VAE. The $+3$ factor arises because the reference frame is replicated at the beginning of the sequence.

To align audio features temporally with these compressed video latents, we apply an audio folding strategy. First, we pad the audio feature sequence $a$ so that its length matches $T{+}3$. Then, we concatenate every four adjacent audio segments along the feature dimension. This converts the original audio tensor of shape $(T{+}3, C)$ into $((T{+}3)/4, 4C)$. The audio feature dimension C extracted by wav2vec is 384. After folding, the feature dimension becomes $4C{=}1536$, which exactly matches the latent channel dimension of Wan2.1-1.3B. As a result, no additional projection layers are required, and the audio features can be integrated directly into the video denoiser via cross-attention.

\subsection{Parts-Specific Reweighting (PSR)}
\label{sec:psr}

Most recent co-speech portrait animation methods follow a similar paradigm: they inject audio features into a pretrained diffusion-based video backbone (e.g., Hallo3 on CogVideo, or EchoMimicV3 and OmniAvatar on Wan2.1). While this allows the model to produce audio-driven portrait motion, it is often insufficient to ensure high-quality synthesis in semantically critical regions such as hands and lips, nor does it guarantee strong audio–visual consistency. This suggests that uniform supervision is not enough for fine-grained spatial control.

\noindent\textbf{Motivation.}
The objective in \cref{eq:ori_flow} implicitly treats all pixels equally. However, perceptually important regions (e.g., lip contours, fingertip articulations) occupy a very small spatial area. Without reweighting, supervision in those regions contributes weakly to the total loss, even though they dominate perceptual quality. Additionally, portrait video frames are highly imbalanced: the dynamic, identity-sensitive foreground contrasts with the static, less important background, making it difficult for the model to learn reliable details in areas like the mouth and hands.%Additionally, portrait video frames exhibit high intra-frame and inter-frame imbalance: the foreground is dynamic and identity-sensitive, while the background is often static and visually unimportant. This imbalance hinders the model from learning reliable local detail in regions like the mouth and hands.

\begin{figure}[htbp]
    \centering  
    \includegraphics[width=\linewidth]{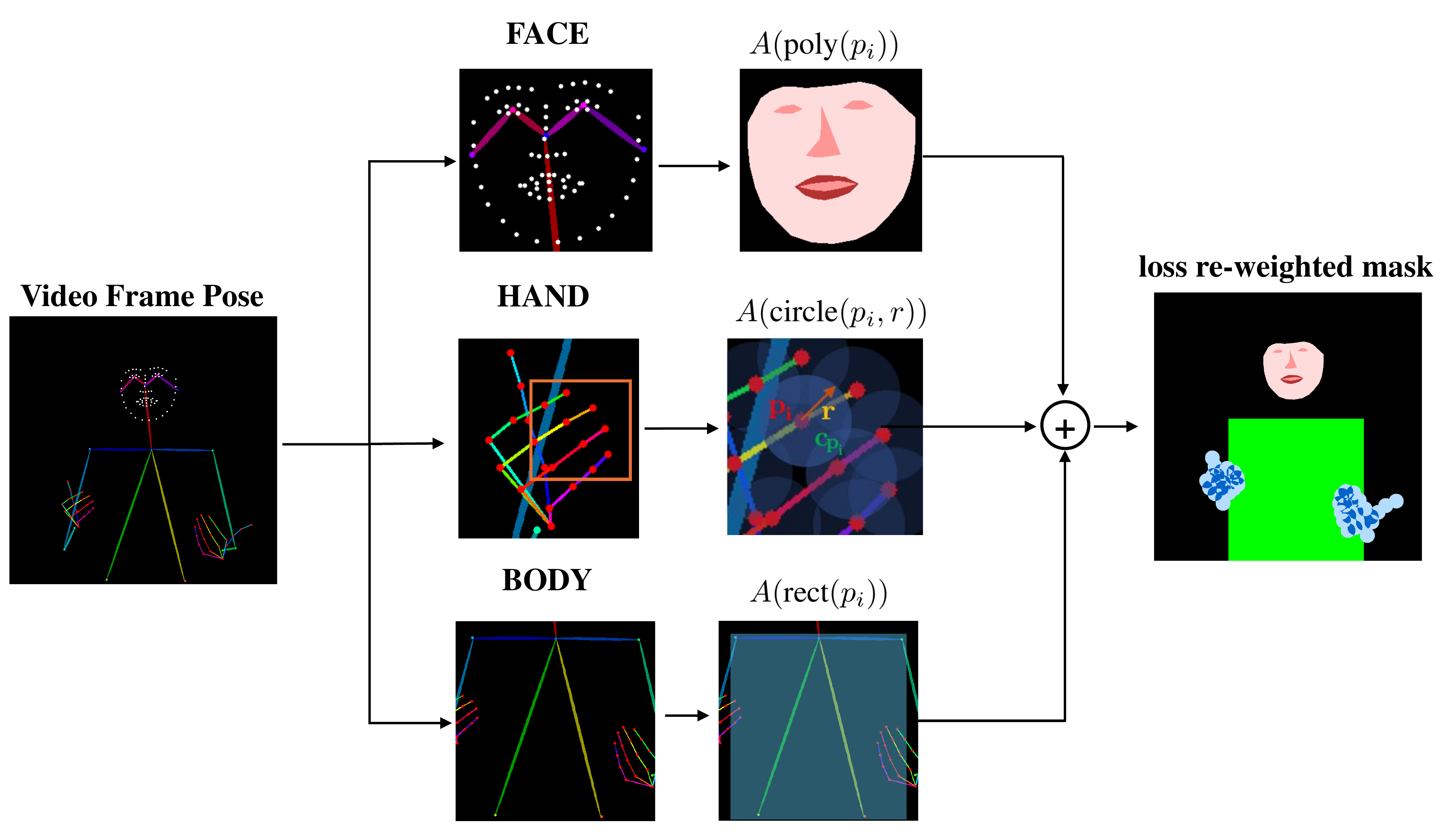}  
    \caption{\textbf{Overview of Parts-Specific Reweighting (PSR).} We obtain pose keypoints and their confidences, construct spatial masks for face, hands, and body, and then combine them into a single reweighting map. Higher weights are assigned to perceptually sensitive regions such as lips and finger gaps.}
    \label{fig:PSR}  
\end{figure}

To resolve this, we introduce \textbf{Parts-Specific Reweighting (PSR)} (Fig.~\ref{fig:PSR}), which adaptively adjusts loss weights across spatial regions using pose confidence and semantic grouping. This encourages the model to allocate more capacity to high-value regions during training.

\noindent\textbf{Awareness regions.}
We divide the portrait into three semantic regions: face, hands, and body. For each frame, we obtain pose keypoints and associated confidence scores using a pose estimator. A keypoint $p_i$ with confidence $c_{p_i}$ is considered reliable if $c_{p_i} > \tau$, where $\tau$ is a fixed threshold. We then derive spatial regions for supervision:

\begin{itemize}
    \item \textbf{Body region.} We construct an axis-aligned bounding rectangle from the min/max $x$ and $y$ coordinates of reliable body keypoints.
    \item \textbf{Hand region.} Around each reliable hand keypoint, we draw a disk of radius $r$ and take the union. Because the gaps between fingers are particularly challenging to model, we additionally emphasize the intersections between adjacent finger disks.
    \item \textbf{Face region.} We expand reliable facial keypoints radially by 10\% in random directions and connect the outermost points to define a polygonal facial mask. Within this mask, we assign different subregions (eyes, nose, mouth, lips, surrounding face skin), with higher weights given to features that are more difficult and more semantically important (e.g., lips).
\end{itemize}

Formally, we define the “awareness areas” as:
\begin{equation}
\resizebox{0.91\linewidth}{!}{$
\begin{aligned}
    S_\text{body} &= S(\text{rect}(p_i)), \quad p_i \in P_\text{body} \,\&\, c_{p_i} > \tau \\[2pt]
    S_\text{eyes} &= S(\text{poly}(p_i)), \quad p_i \in P_\text{eyes} \,\&\, c_{p_i} > \tau \\[2pt]
    S_\text{nose} &= S(\text{poly}(p_i)), \quad p_i \in P_\text{nose} \,\&\, c_{p_i} > \tau \\[2pt]
    S_\text{mouth} &= S(\text{poly}(p_i)), \quad p_i \in P_\text{mouth} \,\&\, c_{p_i} > \tau \\[2pt]
    S_\text{lips} &= S(\text{poly}(p_i)), \quad p_i \in P_\text{lips} \,\&\, c_{p_i} > \tau \\[2pt]
    S_\text{face\_skin} &= S(\text{poly}(p_i)) \setminus 
    \big( S_\text{eyes} \cup S_\text{nose} \cup S_\text{mouth} \cup S_\text{lips} \big), \\
    &\qquad p_i \in P_\text{face} \\[2pt]
    S_\text{finger\_gap} &= 
    \bigcup_{\substack{(p_i, p_j) \in P_\text{fingers} \\ c_{p_i} > \tau, c_{p_j} > \tau}} 
    S(\text{circle}(p_i, r) \cap \text{circle}(p_j, r)) \\[2pt]
    S_\text{hands} &= 
    \left(
        \bigcup_{\substack{p_i \in P_\text{hands} \\ c_{p_i} > \tau}}
        S(\text{circle}(p_i, r))
    \right)
    \setminus S_\text{finger\_gap}.
\end{aligned}
$}
\end{equation}
Here, $S(\cdot)$ denotes the spatial support of the specified geometric region. $\text{circle}(p_i,r)$ is a disk of radius $r$ centered at $p_i$. $P_\text{face}$, $P_\text{eyes}$, etc. correspond to landmark subsets obtained from DWPose~\cite{yang2023dwpose}. $\text{rect}(p_i)$ is the minimum bounding rectangle covering all points $p_i$, and $\text{poly}(p_i)$ is the polygon formed by connecting neighboring points. The operator $\setminus$ indicates set difference.

\noindent\textbf{Loss reweighting.}
We convert these regions into a spatial weight map $M(x)$:
\begin{equation}
M(x) = 
\begin{cases}
        \omega_1, & x \in (S_\text{finger\_gap} \cup S_\text{lips}) \\[1ex]
        \omega_2, & x \in (S_\text{hands} \cup S_\text{eyes} \cup S_\text{nose} \cup S_\text{mouth})  \\[1ex]
        \omega_3, & x \in (S_\text{face\_skin} \cup S_\text{body})  \\[1ex]
        1, & \text{otherwise.}
\end{cases}
\label{eq:psr_weight}
\end{equation}
Here, $\omega_1 > \omega_2 > \omega_3 \geq 1$. Regions that are both challenging and essential for perceptual realism (e.g., lips and finger gaps) are assigned the highest weight $\omega_1$. The next highest weight $\omega_2$ is used for primary semantic subregions of the face (eyes, nose, mouth) and hands. The body and general facial skin, which are easier to reconstruct and less semantically critical, receive a moderate enhancement $\omega_3$. Background regions are left unweighted. \re{More analyses of the PSR weights are provided in the supplementary materials.}

We then apply this map to the training objective. The final training loss becomes:
% \begin{equation}
% \resizebox{0.91\linewidth}{!}{$
% \begin{aligned}
%     \mathcal{L}_{\text{PSR}} 
%     = 
%     Re(M(x)) \cdot
%     \mathbb{E}_{t, z_t, c, \epsilon \sim \mathcal{N}(0, 1)}
%     \left[
%         \| \epsilon_\theta(z_t, t, c) - \epsilon \|_2^2
%     \right].
% \end{aligned}
% $}
% \label{eq:masked_loss}
% \end{equation}
\begin{equation}
\resizebox{0.91\linewidth}{!}{$
\begin{aligned}
    \mathcal{L}_{\text{PSR}} 
    = 
    \mathbb{E}_{t, z_t, c,\epsilon}
    \left[
    \operatorname{mean}
    \left(
    \mathrm{Re}(M(x))
    \odot
    \left|
    \epsilon_\theta(\mathbf z_t,t,c)
    -
    \epsilon
    \right|^2
    \right)
    \right],
\end{aligned}
$}
\label{eq:masked_loss}
\end{equation}
where $\operatorname{mean}$ averages over the channel, temporal, and spatial dimensions, $\mathrm{Re}(\cdot)$ resizes the pixel-space weight map to the spatial resolution of $\mathbf z_t$, and $\odot$ denotes element-wise multiplication.
\re{PSR therefore guides the generator \(G\) to focus more on spatial regions that are critical to human perception in co-speech gesture video generation. This is conceptually similar to BB-TDM\cite{zheng2026BBTDM}, which addresses the limitations of uniform sampling in diffusion models by emphasizing more informative regions or timesteps.}

\subsection{Consistency Classifier Guidance (CCG)}
\label{sec:CCG}

High perceptual quality alone is not sufficient for co-speech gesture generation. The synthesized portrait must also exhibit temporal consistency between speech and visual motion (e.g., head nods, lip activity, hand emphasis). Existing approaches often fail to maintain this alignment across time, leading to audio–visual drift.

\noindent\textbf{Motivation.}
Human speakers exhibit coordinated rhythms: speech prosody, facial articulation, and upper-body gesture are naturally correlated. A diffusion model that conditions only on a reference frame and a long audio segment may not capture these fine-scale temporal correspondences. We address this by introducing an inference-time guidance mechanism that explicitly measures and corrects audio–visual mismatch.

\noindent\textbf{Overview.}
We develop \textbf{Consistency Classifier Guidance (CCG)}, which trains an audio-conditioned latent classifier to discriminate between aligned (real) and misaligned (generated) audio–video pairs. At inference time, this classifier provides gradients that steer the denoising trajectory toward improved audio–motion consistency.

\noindent\textbf{Classifier design.}
To support region-specific control, we train two separate classifiers:
\begin{itemize}
    \item a \textit{\textbf{face classifier}} that focuses on facial regions (e.g., lips);
    \item a \textit{\textbf{non-face classifier}} that focuses on upper-body motion outside the face region.
\end{itemize}
Both classifiers share the same architecture but are trained independently on spatially masked inputs.

\begin{figure}[tbp]
    \centering  
    \includegraphics[width=1\linewidth]{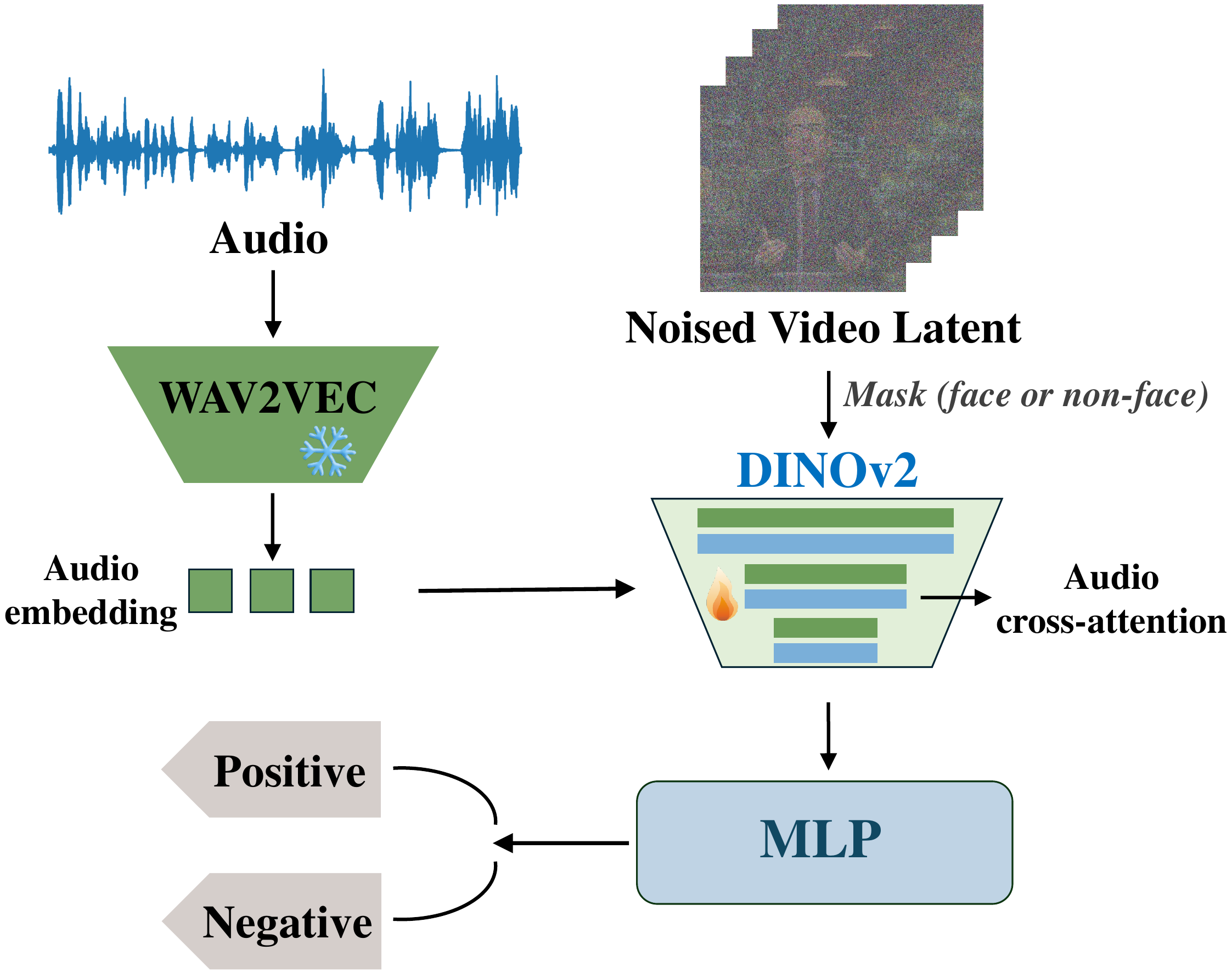}  
    \caption{\textbf{Architecture of the proposed CCG classifier.} We augment a pretrained DINOv2 backbone with audio cross-attention to fuse speech features into the visual latent sequence. A lightweight binary head then predicts whether the audio–video pair is aligned. Separate classifiers are trained for the face and non-face regions.}
    \label{fig:classifier}  
\end{figure}

The classifier architecture is shown in Fig.~\ref{fig:classifier}. The input consists of (i) the current noised video latent and (ii) the corresponding clean audio embedding. Audio features are injected via cross-attention. For the visual branch, we adopt DINOv2~\cite{oquab2024DINOv2} as a lightweight visual backbone. We choose DINOv2 because (1) it provides strong pretrained visual representations, which simplifies downstream training, and (2) it is compact ($\sim$0.03B parameters), so the added computational overhead at inference is minimal. After feature extraction, global pooling is applied, followed by a binary classification head. The classifier outputs a scalar score, in which a higher score indicates stronger audio–visual consistency.

\begin{figure}[htbp]
    \centering
    \includegraphics[width=1\linewidth]{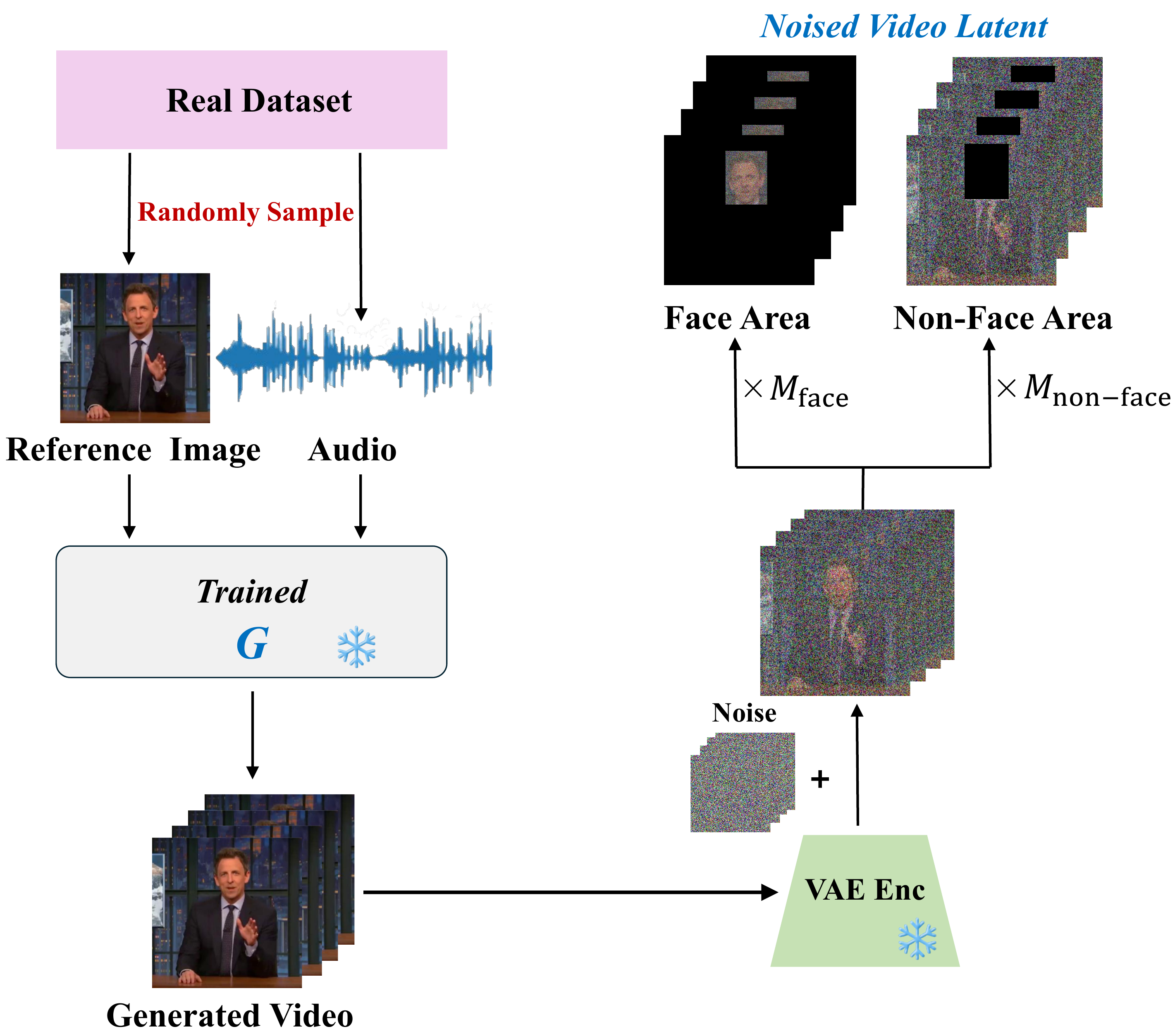}  
    \caption{\textbf{Procedure for constructing negative samples for classifier training.} Negative (misaligned) samples are generated by the trained $G$, while positive samples are taken from real videos. Spatial masking enforces region-specific supervision for the face and non-face classifiers.}
    \label{fig:sample}  
\end{figure}

\noindent\textbf{Training data for CCG.}
To train each classifier, we require both \textit{positive} (aligned) and \textit{negative} (misaligned) examples. Positive samples are taken from real videos in the training set along with their corresponding audio. Negative samples are synthesized using the trained generator $G$ under random sampling settings, as illustrated in Fig.~\ref{fig:sample}. This exposes the classifier to realistic failure modes (e.g., off-beat hand motion, incorrect lip activity).

To ensure that each classifier focuses on its designated spatial region, we apply region masking:
\begin{itemize}
    \item For training the \textit{non-face classifier}, we mask out the face region, emphasizing torso, shoulders, and hands.
    \item For training the \textit{face classifier}, we retain only the face region and mask the rest.
\end{itemize}

We further apply two simple data augmentation strategies to improve robustness: 
(1) The reference frame (identity image) is randomly selected from different points in the source video, rather than fixed to the first frame.  
(2) Audio and video segment lengths are varied, so the classifier learns to judge consistency across different temporal windows.

\noindent\textbf{Classifier loss.}
Let $s$ denote the classifier output (predicted probability that the input pair is consistent) and $y \in \{0,1\}$ the ground truth label. We optimize using binary cross-entropy:
\begin{equation}
       L_{\text{cls}} = - \left[ y \log(s) + (1 - y) \log(1 - s) \right].
\end{equation}

\subsection{Inference Strategy}
\label{sec:inference}

Once trained, the face and non-face classifiers are used only at inference time. At each denoising step, they produce gradients that encourage the generated latent to be consistent with the input speech, separately for facial and non-facial regions. We propose two inference strategies: Sequential Guidance (SG) and Differential Guidance (DG).

\subsubsection{\textbf{Sequential Guidance (SG)}}
\label{sec:SG}

We first sample an initial latent $z_T \sim \mathcal{N}(0, I)$. Let $f_\theta$ denote the numerical solver (denoiser step) conditioned on audio $a$. At each step, we:

1) Predict a latent $\hat{z}_t = f_\theta(\hat{z}^{*}_{t+1}, a, t{+}1)$.

2) Compute a non-face guidance update using the non-face classifier $D_{\text{non-face}}$:
\begin{equation}
\resizebox{0.91\linewidth}{!}{$
    \hat z_{t}^{\text{non-face}} 
    = \hat z_{t} 
    - \lambda_{\text{non-face}} \sigma_t 
    \nabla_{\hat z_{t}} 
    D_{\text{non-face}}(\hat z_{t}, a) 
    * M_{\text{non-face}},
$}
\end{equation}
where $\lambda_{\text{non-face}}$ controls the guidance strength, $\sigma_t$ is the timestep-dependent noise scale, and $M_{\text{non-face}}$ is a spatial mask.

3) Similarly, apply face guidance using $D_{\text{face}}$:
\begin{equation}
   \hat z_{t}^{\text{face}} 
   = \hat z_{t} 
   - \lambda_{\text{face}} \sigma_t 
   \nabla_{\hat z_{t}} 
   D_{\text{face}}(\hat z_{t}, a) 
   * M_{\text{face}}.
\end{equation}

4) Fuse the two guided latents:
\begin{equation}
\begin{aligned}
    \hat z_{t}^{*} 
    = \hat z_{t}^{\text{face}} 
    + \hat z_{t}^{\text{non-face}} 
    - \hat z_t.
    \label{eq:seq}
\end{aligned}
\end{equation}

5) Step to the next time level:
\begin{equation}
    \hat z_{t-1} = f_\theta(\hat z^{*}_{t}, a, t).
    \label{eq:back-infer}
\end{equation}

This process is iterated until $t{=}0$, producing $\hat{z}_0$, which the VAE decoder converts into the final video.

While SG improves regional consistency, we observe interference effects: the face classifier can degrade body–speech synchrony, and the non-face classifier can degrade lip–speech synchrony. This motivates a refined strategy.

\subsubsection{\textbf{Differential Guidance (DG)}}
\label{sec:DG}

% Sequential Guidance (SG) applies both classifiers and then fuses their effects in \cref{eq:seq} and \cref{eq:back-infer}. 
\re{Sequential Guidance (SG) computes guidance updates from both the face and non-face classifiers and directly fuses their effects, as described in \cref{eq:seq} and \cref{eq:back-infer}.
However, we find that these fused updates can introduce cross-region interference: guidance from one region may inadvertently reduce consistency in the other.}

To mitigate this, we propose \textbf{Differential Guidance (DG)}, which explicitly compensates for such interference. Instead of directly applying \cref{eq:back-infer}, DG performs alternating region-specific updates with corrective terms.

First, for the non-face region:
\begin{equation}
\resizebox{0.89\linewidth}{!}{$
  \hat z_{t-1} 
  = f_\theta(\hat z_{t}^{\text{non-face}}, a, t)
  + \lambda_{\text{diff}} \,
  f_\theta(\hat z_{t}^{\text{non-face}}, a, t)
  * M_{\text{face}},
$}
\label{eq:DG_non_face}
\end{equation}
and then, for the face region:
\begin{equation}
\resizebox{0.89\linewidth}{!}{$
  \hat z_{t-2} 
  = f_\theta(\hat z_{t-1}^{\text{face}}, a, t{-}1)
  + \lambda_{\text{diff}} \,
  f_\theta(\hat z_{t-1}^{\text{face}}, a, t{-}1)
  * M_{\text{non-face}}.
$}
\label{eq:DG_face}
\end{equation}

Here, $\lambda_{\text{diff}}$ controls the compensation strength. Intuitively, each classifier first improves its target region, and we then introduce a corrective term that compensates for distortions in the complementary region. This prevents either classifier from degrading areas it was not intended to control.

\re{The full DG inference pipeline with CCG is summarized in Fig.~\ref{fig:inference_ppl} and detailed in Algorithm 1 of supplementary materials. At each diffusion step, we alternate between face-guided and non-face-guided updates, injecting targeted gradients while explicitly correcting cross-region interference. Compared with SG, DG is both more effective (better spatial fidelity and audio–visual synchrony) and more efficient.}
% (\cref{tab:component,tab:inference_time})

\begin{figure}[htbp]
    \centering
    \includegraphics[width=1\linewidth]{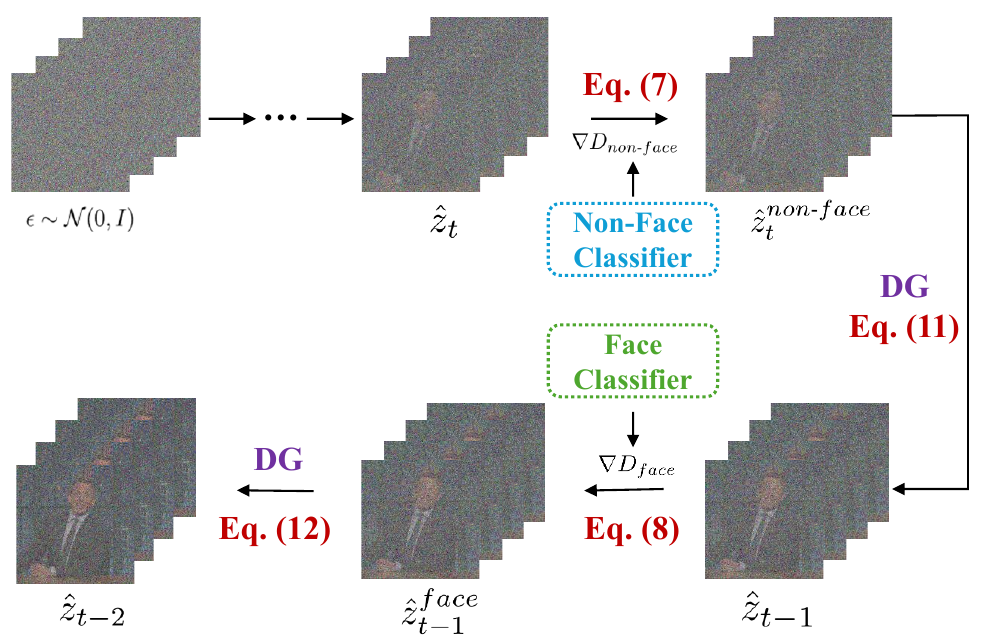}
    \caption{\textbf{Inference pipeline of PAGE with Differential Guidance (DG).} The face and non-face classifiers alternately guide the denoising trajectory. Compensation terms reduce cross-region interference, resulting in improved temporal alignment and local detail compared to naive sequential guidance.}
    \label{fig:inference_ppl}
\end{figure}

\subsection{Chinese Announcer Speech (CAS) Dataset}
\label{sec:dataset}

Prior co-speech gesture generation methods~\cite{liu2022audio,he2024S2G,li2025imae} have primarily been evaluated on English-speaking datasets such as PATS~\cite{ahuja2020PATS}. There is, however, no large-scale benchmark for Mandarin co-speech portrait video synthesis. To address this gap and to enable rigorous evaluation in a linguistically and visually diverse setting, we construct the \textbf{Chinese Announcer Speech (CAS)} dataset.

\cref{tab:dataset_cmp} compares PATS and CAS. CAS offers several advantages:
\begin{itemize}
    \item It contains more video clips and more than twice the total duration of PATS, at a higher average resolution.
    \item It includes over 800 unique speakers, over 200 times
    more than PATS, enhancing robustness to identity variation.
    % \item It includes a substantially larger set of identities (over 800 unique speakers), which improves robustness to identity variation.
    \item It spans a wider variety of scenarios, including indoor/outdoor settings, full-body and half-body framing, standing and seated posture, and a range of expressive speaking styles.
\end{itemize}

\begin{table*}[htbp]  
\centering  
\caption{\textbf{Comparison between the PATS and CAS datasets.} CAS is larger in scale, more diverse in speaking scenarios, and provides high-resolution clips in Mandarin.} 
\label{tab:dataset_cmp}
\begin{tabular}{lcccccccc}  
\toprule  
Dataset & \#Clips & Total Duration & Total Frames & Frames/Clip & \#Identities & Resolution & Full/Half Body & Language \\ 
\midrule  
PATS~\cite{ahuja2020PATS} & 11395 & 145428s (40.40h) & 4362840 & 382.873 & 4 & 1080$\times$1080 & Half & English \\
\textbf{CAS (Ours)} & \textbf{19052} & \textbf{307841s (85.51h)} & \textbf{7684644} & \textbf{403.351} & \textbf{833} & \textbf{1920$\times$1080} & Full and Half & Chinese \\ 
\bottomrule  
\end{tabular}  
\end{table*}

\section{Experiments}
\subsection{Experimental Settings}\label{sec:settings}
\noindent \textbf{Datasets. }
PATS~\cite{ahuja2020PATS} and our proposed CAS are used. Previous works~\cite{liu2022audio,he2024S2G,li2025imae} use the PATS dataset for evaluation, which generally includes four individuals. For fair comparison, we use the same subset as S2G~\cite{he2024S2G}, which comprises four individuals: Jon, Chemistry, Oliver, and Seth. There are a total of 11,395 valid segments (i.e., without cuts or shot changes), each segment lasts 4 to 15 seconds (30 fps), with all samples rescaled to a resolution of 512$\times$512. 90\% of the data is used for training and the remaining 10\% for evaluation. For CAS, there are 19,052 clips containing more than 800 individuals, and the resolution is scaled to 832(H)$\times$480(W).

\noindent\textbf{Evaluation metrics. }
 Following previous works~\cite{he2024S2G,li2025imae}, we evaluate the quality, diversity, and alignment between gestures and speech using four metrics: 1) \textbf{Fréchet Video Distance (FVD)}~\cite{unterthiner2018fvd} to assess the overall quality of gesture videos, computed in feature space using an I3D classifier~\cite{wang2019i3d} pre-trained on Kinetics-400. 2) \textbf{Beat Alignment Score (BAS)}~\cite{li2021bas}, measuring the average similarity between speech and gesture beats, evaluating temporal coherence between speech and gestures. 3) \textbf{Fréchet Gesture Distance (FGD)}~\cite{yoon2020fgd}, measuring the distribution gap between real and generated gestures in feature space; 4) \textbf{Diversity (Div.)}~\cite{liu2022div}, calculating the average feature distance between generated gestures. representing the level of variation.

\subsection{Quantitative Results}
 We compare our method against the following six methods: SDT~\cite{qian2021speech}, S2G~\cite{he2024S2G}, EchoMimicV2~\cite{meng2025echomimicv2}, EchoMimicV3~\cite{meng2025echomimicv3}, OmniAvatar~\cite{gan2025omniavatar}, and Hallo3~\cite{cui2025hallo3}. 
PAGE-DG denotes applying Differential Guidance for the CCG classifier during inference. 
Our model has a similar number of parameters to EchoMimicV3 and OmniAvatar-1.3B (\(\sim\)1.3B). 
SDT cannot be evaluated on CAS because it requires separate training for each identity, which is impractical for CAS's 800+ identities.

The quantitative results reported in~\cref{tab:main_pats} and~\cref{tab:main_cas} show that our method achieves the best performance in quality, alignment, and motion metrics. 
 It exhibits a clear advantage in FVD; for instance, on the CAS dataset, our FVD is 34.76\% lower than Hallo3’s, indicating that the overall quality of our generated videos is closer to the ground truth. On motion-related metrics (FGD and Div), we achieve the best performance; for example, compared with the parameter-matched OmniAvatar-1.3B on CAS, our FGD is reduced by 77.96\%, demonstrating that our method produces realistic and diverse motions. The BAS metric further indicates stronger audio–motion consistency in the generated videos.
Overall, our method can generate realistic videos with high speech gesture consistency under the synergy of the PSR and CCG module.

\begin{table}[htbp]  
\centering  
\caption{\textbf{Quantitative results on PATS.} PAGE-DG means the Differential Guidance (DG) strategy is used for CCG classifier during inference. \textbf{Bold} denotes the best.}  
\begin{tabular}{lcccc}
\toprule
Method & FVD $\downarrow$  & BAS $\uparrow$  & FGD $\downarrow$  & Div.$\uparrow$      \\ 
\midrule
SDT\cite{qian2021speech}    &  294.71 &   0.283  &   11.46  &   13.08   \\
S2G\cite{he2024S2G}   &   268.43   &   0.297    &   2.23  &   13.34     \\
EchoMimicV2\cite{meng2025echomimicv2}  & 257.91 & 0.300 & 3.50 &  13.25    \\
EchoMimicV3\cite{meng2025echomimicv3} &  242.30  & 0.297& 2.79 & 13.28   \\ 
OmniAvatar-1.3B\cite{gan2025omniavatar} & 223.22 & 0.293 & 1.62  &   13.31   \\
OmniAvatar-14B\cite{gan2025omniavatar}           &      215.81    &    0.294   &   1.48    &   13.35      \\
Hallo3\cite{cui2025hallo3}    &   243.75    &   0.297   &   1.43    &   13.42   \\
\midrule
% \textbf{PAGE-SG(Ours) }    &           187.69    &    0.302   &   0.98    &   13.42  \\
\textbf{PAGE-DG(Ours)}     &          \textbf{181.32}    &    \textbf{0.305}   &   \textbf{0.93}    &   \textbf{13.48}   \\
\bottomrule
\end{tabular}
\label{tab:main_pats}
\end{table}

\begin{table}[htbp]  
\centering  
\caption{\textbf{Quantitative results on CAS.} PAGE-DG means the Differential Guidance (DG) strategy is used for CCG classifier during inference. \textbf{Bold} denotes the best.} 
\label{tab:main_cas}
\begin{tabular}{lcccc}
\toprule
Method & FVD $\downarrow$     & BAS $\uparrow$       & FGD $\downarrow$      & Div. $\uparrow$      \\ 
\midrule
S2G\cite{he2024S2G}         &   506.68            &  0.2829         &  1.67             &  12.43    \\
EchoMimicV2\cite{meng2025echomimicv2}     &   1107.65        & 0.2823     & 1.84    &  12.17     \\
EchoMimicV3\cite{meng2025echomimicv3}    &    301.94        &   0.2837        &   1.46        &   12.98     \\
OmniAvatar-1.3B\cite{gan2025omniavatar}        &   384.21     &   0.2833   &  1.18    &  13.11     \\ 
OmniAvatar-14B\cite{gan2025omniavatar}      &    353.65     &  0.2854    &   0.48    &   13.05    \\
Hallo3\cite{cui2025hallo3}  &  296.48    &   0.2864   &   1.12    &   13.11   \\
\midrule
\textbf{PAGE-DG(Ours)}     &          \textbf{193.43}    &    \textbf{0.287}   &   \textbf{0.26}    &   \textbf{13.14}   \\
\bottomrule
\end{tabular}
\end{table}

\subsection{Visualization Results}
\begin{figure}[htbp]
    \centering
    \includegraphics[width=1\linewidth]{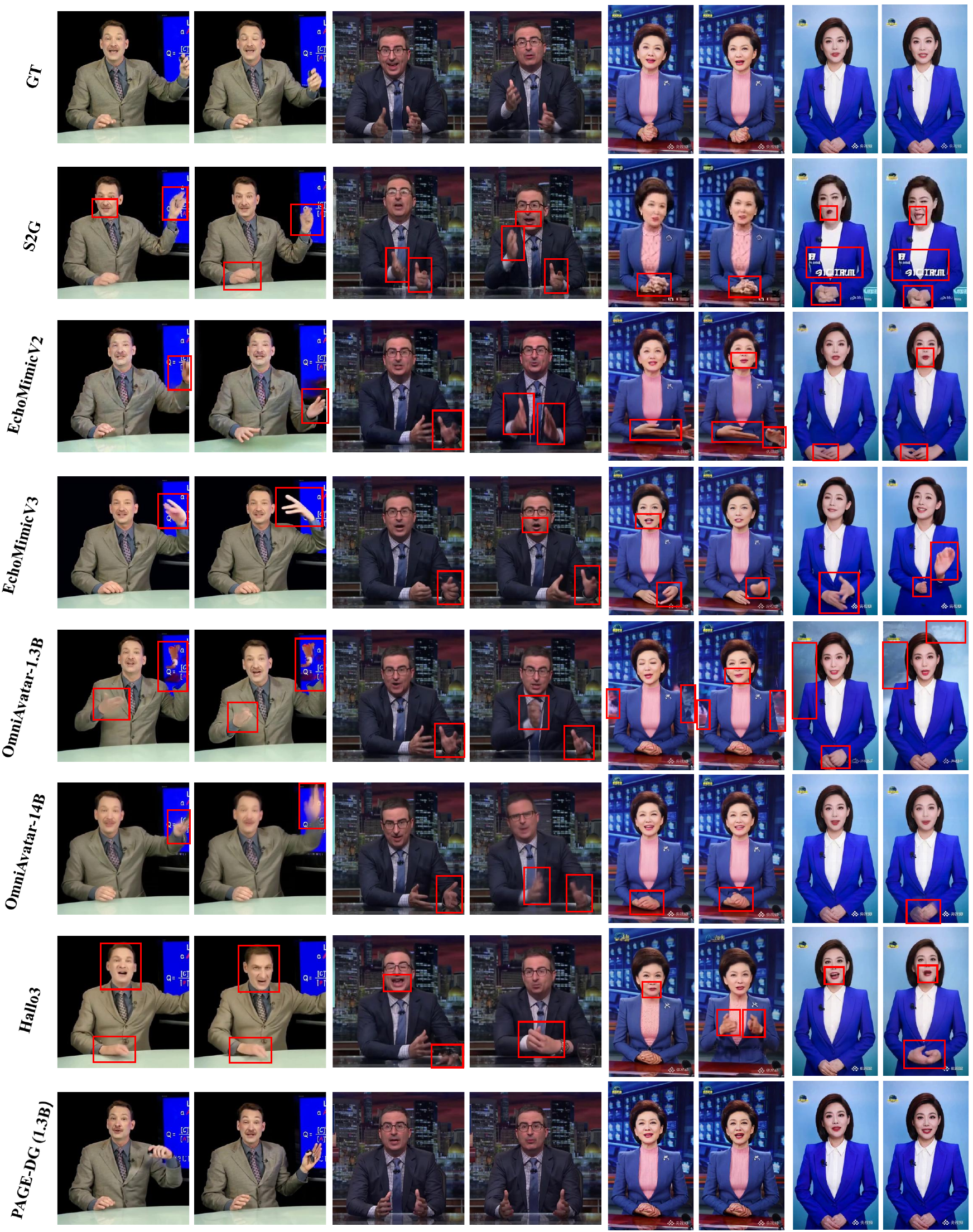}
    \caption{ \textbf{A qualitative comparison with previous work}. We highlight the obvious flaws in earlier methods with red boxes. As shown, previous studies encounter issues like background inconsistencies, hand blurriness, finger distortion, poor face identity preservation and audio-visual misalignment. In contrast, our method can produce high-quality frames.}
    \label{fig:cmp_show}
\end{figure}

%% SDT produces unrealistic shoulder movements.
We conduct a qualitative evaluation of video generation in~\cref{fig:cmp_show}. S2G and OmniAvatar-1.3B produce videos with comparatively poor visual quality, particularly in portrait and clothing details. EchoMimicV2 and EchoMimicV3 demonstrate improvements in speech-motion synchronization, but EchoMimicV2 frequently exhibits artifacts in hand regions, while EchoMimicV3 struggles with maintaining facial identity consistency. OmniAvatar-14B offers superior visual quality, yet suffers from limited motion diversity and frequent mismatches between speech and motion. Hallo3 exhibits noticeable artifacts in both facial and hand regions, resulting in unsatisfactory local generation quality.
In contrast, our method generates high-quality videos, especially in the face and hand regions, significantly enhancing the consistency between speech and motion by using a carefully trained CCG classifier for guidance during inference.

\subsection{User Study}
Objective metrics correlate imperfectly with human perception, and audio-driven video generation is often interactive, making subjective human perception more compelling than objective metrics. To evaluate our method's visual performance, we conduct a user study to assess the quality of videos generated by various methods. For each method, we choose 30 generated videos from the PATS and CAS test sets for comparison. We invite 20 participants for evaluation. Participants choose the best video based on four criteria: 1) \textbf{Realism}, 2) \textbf{Diversity}, 3) \textbf{Synchrony}, and 4) \textbf{Overall Quality}. Diversity focuses on the diversity of co-speech gestures. Synchrony measures the alignment of speech and gesture. Participants are instructed to disregard irrelevant factors like texture and facial expressions during motion-related assessments. We calculate the preferred percentage for each criterion independently. 
The results, shown in~\cref{tab:user-study}, demonstrate that our method significantly outperforms others on all criteria, particularly in Realism, Diversity, and Overall Quality, indicating that our framework generates superior co-speech gesture videos while effectively balancing motion and visual effects. For Synchrony, we find that evaluating speech and motion, including lip consistency, is subjective and therefore prone to randomness among different participants. Even so, our method significantly outperforms other methods on Synchrony.

\begin{table}[htbp]
\centering
\caption{\label{tab:user-study} \textbf{User preferred percentage for each method.} Test videos from PATS and CAS.
}
\resizebox{\linewidth}{!}{ 
\begin{tabular}{lcccc}
\toprule
 Method   & {Realism $\uparrow$} & {Diversity $\uparrow$} & {Synchrony $\uparrow$} & {Overall Quality $\uparrow$} \\
\midrule
S2G~\cite{he2024S2G}    &  0.0\%   &   0.0\%  &  1.0\%   &   0.0\%   \\
EchoMimicV2~\cite{meng2025echomimicv2}  &  4.5\%    &    7.5\% &  20.5\%   &   1.5\%   \\
EchoMimicV3~\cite{meng2025echomimicv3}        &  0.5\%   &   3.5\%  &  7.5\%   &    3.5\%   \\
OmniAvatar-1.3B~\cite{gan2025omniavatar}  &  0.0\%   &   0.0\%  &  0.5\%   &    0.0\%   \\
OmniAvatar-14B~\cite{gan2025omniavatar}  & 16.0\%   &   3.0\%  &  6.0\%   &    17.0\%  \\
Hallo3~\cite{cui2025hallo3}  & 5.0\%   &   9.5\%  &  10.33\%   &    5.5\%  \\
\midrule
\textbf{PAGE-DG(Ours)}      & \textbf{74.0\%} & \textbf{76.5\%}&  \textbf{54.17\%} & \textbf{72.5\%} \\ 
\bottomrule
\end{tabular}
}
\end{table}

\subsection{Ablation Study}

We perform ablation experiments on the PATS test set to evaluate core framework components and hyperparameters.

\noindent \textbf{Core Components. }
PAGE comprises PSR and CCG, with DG and SG as the proposed and baseline inference strategies, respectively.~\cref{tab:component} presents the ablation results and verifies the contribution of each module.
The results show that omitting PSR leads to an increase in the video quality metric FVD from 181.32 to 189.04, while the Div metric decreases from 13.48 to 13.44. The absence of Classifier Guidance has the most substantial impact, as reflected across all four metrics. For example, the FVD increases from 181.32 to 212.46, while the FGD rises from 0.93 to 1.42. Both BAS and Div also exhibit substantial changes. These results highlight the critical role of our Classifier Guidance design. Furthermore, we observe that the DG consistently outperforms the SG across all four metrics. As discussed in Sec.~\ref{sec:DG}, this can be attributed to the negative interference between facial and non-facial classifiers in the SG, whereas the DG effectively eliminates this adverse effect through a compensation mechanism.

\begin{table}[htbp]  
\centering  
\caption{\label{tab:component}\textbf{Ablation of core components.}} 
\resizebox{\linewidth}{!}{
\begin{tabular}{lcccc}  
\toprule  
Name                         & FVD $\downarrow$ & BAS $\uparrow$  & FGD $\downarrow$  & Div. $\uparrow$  \\ \midrule  
Wan-1.3B-Baseline       & 221.87         & 0.289         & 1.51            & 13.28          \\
% W PSR(Mimic)          & 214.73         & 0.292         & 1.45             & 13.29          \\
w/ PSR          & 212.46          & 0.293          & 1.42             & 13.33           \\
w/ SG          & 194.53          & 0.300          & 1.05                & 13.39          \\ 
w/ DG          & 189.04          & 0.303         & 1.01                  & 13.44           \\ 
w/ PSR+SG           & 187.69          & 0.302         & 0.98                  & 13.42           \\  
w/ PSR+DG          & \textbf{181.32}   & \textbf{0.305}    & \textbf{0.93}           & \textbf{13.48} \\ \bottomrule  
\end{tabular}  
}
\end{table}

\noindent \textbf{Guidance Step Rate. }

\re{When the \textit{Guidance Step Rate} is below 100\%, e.g., 80\%, our guidance strategy
% (\cref{alg:inference})
is applied to 80\% of early inference steps, while the remaining 20\% use standard denoising update.} \cref{tab:guide_step} compares PAGE-SG (Sequential Guidance) and PAGE-DG (Differential Guidance) across different guidance rates for CCG-based inference. The first number in each pair is PAGE-SG, the second is PAGE-DG. Results indicate that higher guidance rates enhance performance, confirming the adaptability of our CCG classifier to diverse data distributions.

\begin{table}[htbp]  
\centering  
\caption{\textbf{Ablation of guidance step rate}. In each number pair, the former is PAGE-SG and the latter is PAGE-DG.
}   
\label{tab:guide_step}
\resizebox{\linewidth}{!}{
\begin{tabular}{cccccc}
\toprule
Guidance Step Rate & FVD $\downarrow$     & BAS $\uparrow$       & FGD $\downarrow$       & Div. $\uparrow$     \\ 
\midrule
0\%             &     212.46/\;212.46              &  0.293/\;0.293         &     1.42/\;1.42              &    13.33/\;13.33   \\
20\%             &      203.74/\;198.53         &     0.295/\;0.299        &     1.22/\;1.16           &     13.35/\;13.36     \\
40\%             &     190.13/\;189.61   & 0.298/\;0.304    & 1.05/\;0.98     & 13.40/\;13.42   \\
60\%      &       189.76/\;185.98       &  0.297/\;0.302       &    1.02/\;0.98        &   13.39/\;13.45   \\
80\%    &   189.13/\;185.27        &  0.301/\;0.303         &  \textbf{0.97}/\;0.95     &   13.40/\;13.47      \\
100\%    &  \textbf{187.69/\;181.32}        & \textbf{ 0.302/\;0.305}        &  0.98/\;\textbf{0.93}     &    \textbf{13.42/\;13.48}
\\ 
\bottomrule
\end{tabular}
}
\end{table}

\noindent \textbf{CCG Classifier. }
For CCG, we evaluated several classifier architectures as summarized in~\cref{tab:ablation_classifier}. Vit-S-PAGE is a custom model based on Vit-S, while Wan2.1 (2 Blocks) and Wan2.1 (30 Blocks) refer to variants of Wan2.1 utilizing 2 and 30 Dit blocks, respectively. We report both classification accuracy and the FVD, the latter being more relevant to our task as it assesses the quality of generated videos guided by the CCG classifier. The results indicate that increasing model size, as in Wan2.1 (30 Blocks), does not necessarily lead to better performance. DINOv2 was ultimately selected, as it offers a favorable balance between parameters and generation quality.
\begin{table}[htbp]  
\centering  
\caption{\label{tab:ablation_classifier}\textbf{Ablation of CCG Classifier.}} 
\resizebox{\linewidth}{!}{
\begin{tabular}{lccc}  
\toprule  
Name  & \#Param. & Accuracy $\uparrow$  & FVD $\downarrow$  \\ 
\midrule  
Vit-S-PAGE\cite{dosovitskiy2021Vit}  & 0.03B    & 0.731    & 210.05    \\
Wan2.1(2 Blocks)   & 0.09B   & 0.975  & 192.38 \\
Wan2.1(30 Blocks)    & 1.3B   & 0.987   & 189.54 \\ 
DINO-v3-based\cite{simeoni2025dinov3}  & 0.05B  & \textbf{0.992}      & 186.76          \\ 
DINO-v2-based\cite{oquab2024DINOv2}   & 0.03B          & 0.979   & \textbf{181.32}    \\  
\bottomrule  
\end{tabular}  
}
\end{table}

\noindent\textbf{Audio injection.}
As shown in \cref{tab:audio_inject}, \textit{Normal Injection} feeds the raw audio into wav2vec without prepending the audio corresponding to the first frame, producing features of size $(T, 384)$, where $T$ is the number of training frames. A linear layer then projects 384 to 1536 to match Wan2.1. However, in the Wan2.1-based generative framework, the reference image is triplicated before VAE encoding, causing the temporal dimension to become $T+3$ rather than $T$. Thus, \textit{Latent Alignment} replicates the first-frame audio three times and prepends the copies to the sequence. Wav2vec outputs audio features of size $(T+3, 384)$; concatenating every four consecutive frames yields the final audio feature map of size $((T+3)/4, 1536)$. The resulting temporal length $((T+3)/4)$ and channel dimension (1536) match the visual features after Wan2.1 VAE downsampling, removing the need for a linear projection and enabling tighter audio–visual alignment.

\begin{table}[htbp]  
\centering  
\caption{\textbf{Ablation of audio injection.}} 
\label{tab:audio_inject}
\resizebox{\linewidth}{!}{
\begin{tabular}{lccccc}  
\toprule  
Name  & FVD $\downarrow$ & BAS $\uparrow$  & FGD $\downarrow$  & Div. $\uparrow$  \\ 
\midrule  
Normal injection          & 214.73         & 0.292         & 1.45             & 13.29          \\
Latent alignment(ours)         & \textbf{212.46}          & \textbf{0.293}          & \textbf{1.42}             & \textbf{13.33}           \\   
\bottomrule 
\end{tabular}  
}
\end{table}

\noindent \textbf{Guidance Weight. }
 Tables~\ref{tab:face_nonface_guidance_weight} and~\ref{tab:DG_guidance_weight} present the quantitative results of the guidance weight $\lambda_{non\text{-}face}$, $\lambda_{face}$, and differential guidance weight $\lambda_{diff}$ during inference. The final weight combination is set to $(\lambda_{non\text{-}face}=1, \lambda_{face}=1, \lambda_{diff}=1)$, achieving optimal performance.
\begin{table}[htbp]  
\centering  
\caption{\textbf{Ablation of two classifiers’ guidance weights.}}
\label{tab:face_nonface_guidance_weight}
\resizebox{\linewidth}{!}{
\begin{tabular}{lccccc}
\toprule
$\lambda_{face}$  & $\lambda_{non\_face}$  & FVD $\downarrow$     & BAS $\uparrow$       & FGD $\downarrow$        & Div. $\uparrow$      \\ 
\midrule
2             &       1      &     194.07             &      0.301         &      1.14        &   13.4    \\
0.5            &    1      & 203.14       & 0.296        & 1.2      & 13.35       \\
1             &     1             &   \textbf{187.69}               &      \textbf{0.302}            &      \textbf{0.98}               &   \textbf{13.42}    \\
1            &      0.5            &       201.56          &        0.297            &     1.17                &    13.38   \\
1            &      2         &         190.63        &      \textbf{0.302}      &     1.08     &               13.4         \\ 
\bottomrule
\end{tabular}
}
\end{table}

\begin{table}[htbp]  
\centering  
\caption{ \textbf{Ablation of differential guidance weight.} } 
\label{tab:DG_guidance_weight}
\resizebox{0.95\linewidth}{!}{
\begin{tabular}{lcccc}
\toprule
$\lambda_{diff}$ & FVD $\downarrow$     & BAS $\uparrow$       & FGD $\downarrow$       & Div. $\uparrow$      \\ 
\midrule
0.5            &        198.24         &   0.298          &    1.12                &   13.4     \\
1            &        \textbf{181.32}        &     \textbf{0.305}          &     \textbf{0.93}                &   \textbf{13.48}    \\
2            &   187.98    & 0.302   & 0.99    & 13.43       \\  
\bottomrule
\end{tabular}
}
\end{table}

\subsection{Inference Time. } 
\re{We measured the average inference time for each method by generating several 10-second video clips, as shown in~\cref{tab:inference_time}. Our method has a similar inference speed to models with comparable parameters(\(\sim\)1.3B), such as EchoMimicV3 and OmniAvatar-1.3B, since PSR is only used during training and does not affect inference. The proposed CCG module, built upon the lightweight DINOv2 model, adds minimal overhead: removing it reduces PAGE-DG’s inference time from 140.6s to 139.4s (only 0.9\%).
Compared with PAGE-SG, PAGE-DG is both more efficient and more effective, as it requires less computational overhead and eliminates the mutual interference between face and non-face regions.}

\begin{table}[htbp]  
\centering  
\caption{\textbf{Inference time} refers to generating a \textit{10s} video, all experiments conducted under identical conditions.}
\label{tab:inference_time}
\resizebox{\linewidth}{!}{
\begin{tabular}{lc|lc}
\toprule
Method & Inference Time(s) & Method & Inference Time(s)\\
\midrule
S2G\cite{he2024S2G} & 23.8 & Hallo3\cite{cui2025hallo3}& 3569.2\\
EchoMimicV3(1.3B)\cite{meng2025echomimicv3} & 140.1 & \textbf{PAGE w/o CCG} & 139.4 \\
OmniAvatar-1.3B\cite{gan2025omniavatar} & 138.4 & \textbf{PAGE-SG} & 142.2\\
OmniAvatar-14B\cite{gan2025omniavatar} & 787.0 &  \textbf{PAGE-DG} & 140.6 \\
\bottomrule
\end{tabular}
}
\end{table}

\section{Conclusion}
We introduced PAGE, a unified diffusion-based framework for co-speech gesture portrait video generation. PAGE enhances spatial fidelity through Parts-Specific Reweighting (PSR) and improves temporal audio–visual alignment using Consistency Classifier Guidance (CCG) and Differential Guidance (DG).  
In addition, we established the Chinese Announcer Speech (CAS) dataset, the first large-scale Chinese benchmark for co-speech gesture generation.  
Extensive evaluations demonstrate that PAGE achieves state-of-the-art performance on both PATS and CAS, producing high-quality, expressive, and synchronized talking videos.

\clearpage

% \section*{Supplementary Material}
% \section*{\textbf{Supplementary Material}}

\twocolumn[
\vspace*{1em}
\begin{center}
    \textbf{\Large Supplementary Material}
\end{center}
\vspace{2em}
]

\renewcommand{\thesection}{\Alph{section}}
\setcounter{section}{0}

\section{Overview}
This supplementary material provides additional details and analyses, including:
\begin{itemize}
    \item implementation details of PAGE;
    \item adaptation to different diffusion frameworks;
    \item inference pipeline of Differential Guidance (DG);
    \item PSR weight setting analysis;
    \item CAS dataset construction and visualization;
    \item limitations and future work.
\end{itemize}
We also provide supplementary videos for qualitative comparison.

\section{Additional Implementation Details}\label{sec:supp_details}

When training the backbone $G$ using our PSR, we use DWpose~\cite{yang2023dwpose} to extract the pose sequences. For the reference image, we extract semantic image embeddings using CLIP and visual latent features using the VAE encoder of Wan2.1.
%The visual encoder of Wan2.1 is used to encode CLIP embeddings of reference images. 
% We initialize the model using a pretrained video generation model~\cite{wan2025wan}. 
During training, the resolution H$\times$W of videos is 512$\times$512 on PATS (832$\times$480 on CAS). The sampled frame length is fixed at 81 frames, and the batch size is 8. We use the AdamW optimizer with a learning rate of 1e-5 to optimize the network, and the number of training diffusion timesteps is set to 1000. The keypoint confidence score threshold $\tau$ is set to 0.8, and the radius $r$ is 30. $\omega_1$, $\omega_2$, $\omega_3$ in Eq. (4) are set to 10, 5, 2, respectively.

In the CCG classifier training, the positive audio--visual samples come from PATS and CAS, while the negative samples are generated by the trained $G$ using randomly selected denoising steps among the 25 inference steps, forming one-to-one matched positive and negative pairs. To further augment the training set, we employ the data augmentation strategy described in \textit{Consistency Classifier Guidance(CCG)} section, where the lengths of the audio and video are uniformly sampled among 30, 60, 90, and 120 frames. With these augmentation strategies, we obtain more than 50,000 pairs of audio--visual samples. We use MediaPipe~\cite{lugaresi2019mediapipe} to obtain facial boxes. Both the non-face classifier $D_{non\text{-}face}$ and face classifier $D_{face}$ use a learning rate of 1e-4, with a batch size of 16 and 100k training steps. A warmup strategy is applied for the first 5000 steps, and the weights of the encoder part of the two classifiers are the same with the trained $G$. 
%Experiments are conducted on 8 NVIDIA H20 GPUs.

During the final inference, the Classifier Guidance is applied in all 25 inference steps. We set $\lambda_{face}$, $\lambda_{non\text{-}face}$ and $\lambda_{diff}$ to 1. The gradient returned by the CCG classifier naturally represents the difference between the inference result and the GT, and directly adding it serves as a refinement. 

\newpage

\section{Adaptation to Different Diffusion Frameworks}

PAGE is designed as a general optimization framework of diffusion-based co-speech gesture portrait video generation at two levels: first, at the training stage, we improve the \textbf{training objective} so that the model explicitly focuses on perceptually important regions such as lips, eyes, hands, and finger gaps; second, at the inference stage, we optimize the \textbf{denoising trajectory} and directly correct audio-motion inconsistencies through audio--visual consistency guidance. Therefore, PAGE is not simply a reuse of an existing diffusion backbone. Instead, it is a parts-aware training and inference framework that can be transferred to different diffusion architectures.

To further demonstrate the generality of PAGE, we have added experiments across different diffusion model architectures. 
\cref{tab:ablation_wan_dit} reports the results on the Wan2.1 video generation model~\cite{wan2025wan}, which adopts a DiT-based backbone and follows the Flow Matching (FM)~\cite{lipman2023flow} training framework. In addition, we have added experiments on Stable Video Diffusion (SVD)~\cite{blattmann2023SVD} in \cref{tab:ablation_svd_unet}, which uses a U-Net-based backbone and is trained under the DDPM~\cite{ho2020denoising} framework.
The results show that our proposed PSR, CCG, and DG also bring consistent improvements under the SVD setting. This indicates that PAGE is not an engineering combination tied to a specific backbone or diffusion training framework, but a general optimization of the training objective and inference trajectory of diffusion models.

\begin{table}[htbp]  
\centering  
\caption{\textbf{Ablation of the proposed PSR, CCG, and DG modules on Wan.}} 
\label{tab:ablation_wan_dit}
\resizebox{1\linewidth}{!}{
\begin{tabular}{ccccccc}  
\toprule  
\makecell{Wan2.1\\\scriptsize{(DiT+FM)}} & PSR &  CCG+DG  & FVD $\downarrow$ & BAS $\uparrow$  & FGD $\downarrow$  & Div. $\uparrow$  \\ 
\midrule  
\cmark & \xmarkg &  \xmarkg     & 221.87         & 0.289         & 1.51            & 13.28          \\
\cmark   & \cmark  &  \xmarkg    & 212.46          & 0.293         & 1.42     & 13.33    \\
\cmark    & \xmarkg  &  \cmark   & 189.04     & 0.303   & 1.01     & 13.44    \\ 
\cmark    & \cmark  &  \cmark    & \textbf{181.32}   & \textbf{0.305}    & \textbf{0.93}  & \textbf{13.48} \\ \bottomrule  
\end{tabular}  
}
\end{table}

\begin{table}[htbp]  
\centering  
\caption{\label{tab:ablation_svd_unet}\textbf{Ablation of the proposed PSR, CCG, and DG modules on SVD.}} 
\resizebox{1\linewidth}{!}{
\begin{tabular}{ccccccc}  
\toprule  
\makecell{SVD\\\scriptsize{(U-Net+DDPM)}} & PSR &  CCG+DG  & FVD $\downarrow$ & BAS $\uparrow$  & FGD $\downarrow$  & Div. $\uparrow$  \\ 
\midrule  
\cmark & \xmarkg &  \xmarkg     & 260.14         & 0.281         & 4.07            & 13.16          \\
\cmark   & \cmark  &  \xmarkg    & 253.06          & 0.283        & 3.61     & 13.18    \\
\cmark    & \xmarkg  &  \cmark   & 244.23     & 0.288   & 3.32     & 13.23    \\ 
\cmark    & \cmark  &  \cmark    & \textbf{241.07}   & \textbf{0.291}    & \textbf{3.05}  & \textbf{13.24} \\ \bottomrule  
\end{tabular}  
}
\end{table}

\newpage
\section{Inference pipeline of Differential Guidance}\label{sec:ppl_infer}
The inference process for CCG (Consistency Classifier Guidance) with DG (Differential Guidance) is illustrated in~\cref{alg:DG_inference}. At each diffusion step, face and non-face classifiers alternately provide gradient-based guidance, with additional compensation terms introduced to mitigate adverse interactions between the two regions. Compared with the naive SG (Sequential Guidance) approach, our DG strategy offers both improved efficiency and enhanced generation performance.

\begin{algorithm}[htbp]
\caption{Differential Guidance Inference}
\label{alg:DG_inference}
\small
\renewcommand{\baselinestretch}{1.35}\selectfont
\begin{algorithmic}[1]
    \Require reference image $\mathbf{I}_{\text{ref}}$, driving audio $\mathbf{a}$, trained generator $G_{\mathrm{PSR}}$, face classifier $D_{\text{face}}$, non-face classifier $D_{\text{non-face}}$, timesteps $t_N > \cdots > t_0$, ODE solver $f_\theta$, noise schedule $\alpha(t), \sigma(t)$, face mask $M_{\text{face}}$, non-face mask $M_{\text{non-face}}$, guidance weights $\lambda_{\text{face}}, \lambda_{\text{non-face}}, \lambda_{\text{diff}}$, decoder $\mathbf{D}$
    \Ensure generated co-speech gesture video $\mathbf{v}$

    \State Initialize latent: $\hat{\mathbf{z}}_{t_N} \sim \mathcal{N}(\mathbf{0}, \mathbf{I})$
    
    \For{$n \gets N$ down to $1$}
        \If{$n \bmod 2 = 0$} 
            \State \hspace{1em} \textit{// non-face guidance step}
            \State $
            \begin{aligned}[t]
            \hat{\mathbf{z}}^{\text{non-face}}_{t_n} \gets {}&
              \hat{\mathbf{z}}_{t_n}
              - \lambda_{\text{non-face}}\,\sigma(t_n)\,
              \nabla_{\hat{\mathbf{z}}_{t_n}} D_{\text{non-face}}(\hat{\mathbf{z}}_{t_n}, \mathbf{a}) \\
              & * M_{\text{non-face}}
            \end{aligned}
            $
            \State $\mathbf{b} \gets f_\theta(\hat{\mathbf{z}}^{\text{non-face}}_{t_n}, \mathbf{a}, t_n)$
            \State $\hat{\mathbf{z}}_{t_{n-1}} \gets \mathbf{b} + \lambda_{\text{diff}} \, \mathbf{b} * M_{\text{face}}$
        \Else 
            \State \hspace{1em} \textit{// face guidance step}
            \State $\hat{\mathbf{z}}^{\text{face}}_{t_n} \gets \hat{\mathbf{z}}_{t_n}
            - \lambda_{\text{face}} \, \sigma(t_n) \,
            \nabla_{\hat{\mathbf{z}}_{t_n}} D_{\text{face}}(\hat{\mathbf{z}}_{t_n}, \mathbf{a}) * M_{\text{face}}$
            \State $\mathbf{b} \gets f_\theta(\hat{\mathbf{z}}^{\text{face}}_{t_n}, \mathbf{a}, t_n)$
            \State $\hat{\mathbf{z}}_{t_{n-1}} \gets \mathbf{b} + \lambda_{\text{diff}} \, \mathbf{b} * M_{\text{non-face}}$
        \EndIf
    \EndFor

    \State $\mathbf{v} \gets \mathbf{D}(\hat{\mathbf{z}}_{t_0})$
\end{algorithmic}
\end{algorithm}

\section{PSR Weight Settings Analysis}

To examine whether PSR depends on one specific manually selected weight configuration, we conduct experiments in~\cref{tab:psr_weight_select} with multiple fixed weight settings corresponding to different reweighting strengths:

\begin{itemize}
    \item \textbf{Uniform weighting}: $(\omega_1,\omega_2,\omega_3)=(1,1,1)$, which applies no region reweighting and serves as the no-PSR baseline;
    \item \textbf{Weak reweighting}: $(\omega_1,\omega_2,\omega_3)=(2.5,1.5,1)$, which applies mild region enhancement;
    \item \textbf{Default setting}: $(\omega_1,\omega_2,\omega_3)=(10,5,2)$, which is the setting used in our main experiments;
    \item \textbf{Strong reweighting}: $(\omega_1,\omega_2,\omega_3)=(24,10,4)$, which is used to test the stability under stronger region enhancement.
\end{itemize}

\begin{table}[htbp]  
\centering  
\caption{\label{tab:psr_weight_select}\textbf{Ablation with different PSR weight settings.}} 
\resizebox{1\linewidth}{!}{
\begin{tabular}{ccccc}  
\toprule  
$(\omega_1,\omega_2,\omega_3)$ & FVD $\downarrow$ & BAS $\uparrow$  & FGD $\downarrow$  & Div. $\uparrow$  \\ 
\midrule  
(1,1,1) & 221.87 &  0.289   &   1.51    & 13.28  \\
(2.5,1.5,1) & 213.14  & 0.291  &  1.45    & \textbf{13.36}    \\
(10,5,2) & 212.46  & 0.293  & \textbf{1.42}   & 13.33 \\ 
(24,10,4) &  \textbf{210.96}  & 0.293  & 1.43   & 13.29 \\ 
Adaptive Learning & 211.37 & \textbf{0.297}  & 1.44   & 13.31 \\ 
 \bottomrule  
\end{tabular}  
}
\end{table}

The results show that uniform weighting performs clearly worse, indicating that applying the same supervision strength to all regions leads to under-optimization of small but perceptually important regions, such as lips, hands, and finger gaps. In contrast, the other PSR weight settings consistently improve the performance, showing that region-aware supervision itself is effective. Meanwhile, the performance differences among the weak, default, and strong PSR settings are relatively small. This suggests that PSR is not sensitive to the exact weight values within a reasonable range, and its effectiveness does not rely on a fragile manually selected weight configuration.

Furthermore, we have also added an adaptive weighting variant, where the model learns the corresponding region weights during training. The results (\cref{tab:psr_weight_select}) show that this adaptive variant does not provide stable and significant improvements over fixed PSR, while it may introduce additional model complexity and training instability. Therefore, fixed weighting is a simpler and more reproducible choice.
For the adaptive learning variant, we further inspect the final learned weights, which converged to \((\omega_1,\omega_2,\omega_3)=(13.16, 9.42, 1.89)\), preserving the same hierarchical order as our fixed PSR design.

\section{CAS Dataset}

\subsection{\textbf{Dataset construction.}}

\begin{figure*}[htbp]
  \centering
   \includegraphics[width=1\linewidth]{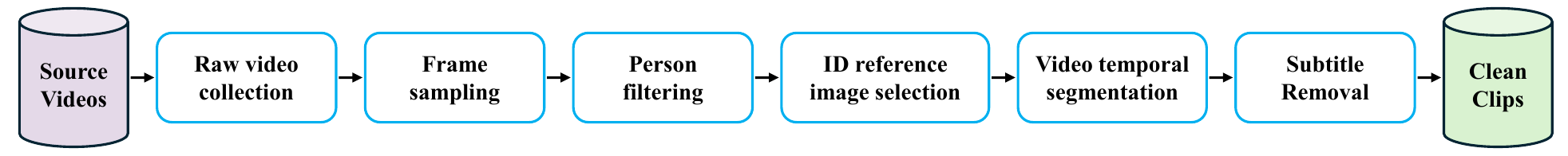}
   \caption{\textbf{CAS dataset construction pipeline.}}
   \label{fig:cas_dataset_ppl}
\end{figure*}

To obtain high-quality single-speaker clips and remove unsuitable content (e.g., shots with no person, tiny faces, scene cuts, or strong occlusions), we design a multi-stage processing pipeline. The overall process is illustrated in~\cref{fig:cas_dataset_ppl}, and the steps are described below.

\begin{enumerate}
    \item \textbf{Raw video collection. }
    The raw videos were mainly collected from publicly accessible media platforms, including news websites, television-station websites, and online video platforms. These videos cover diverse scenarios, including indoor and outdoor settings, full-body and half-body framing, standing and seated postures, and a range of expressive speaking styles.

    During data collection, we applied basic quality requirements to the raw videos: \textit{\textbf{video bitrate no lower than 2 Mbps, resolution no lower than \(1920\times1080\), frame rate no lower than 25 fps, and audio bitrate no lower than 128 kbps.}}

    To obtain clips in which the main subject consistently remains a single person throughout the video (\cref{fig:cas_good_clip}), we designed the following processing pipeline. This pipeline filters out unsuitable raw-video cases, including speaker or identity switching, background-only frames, very small person regions, multiple visible persons, strong occlusions, and hard scene cuts (\cref{fig:cas_bad_clip_1,fig:cas_bad_clip_2}).
    \begin{figure}[htbp]
      \centering
       \includegraphics[width=1\linewidth,page=1]{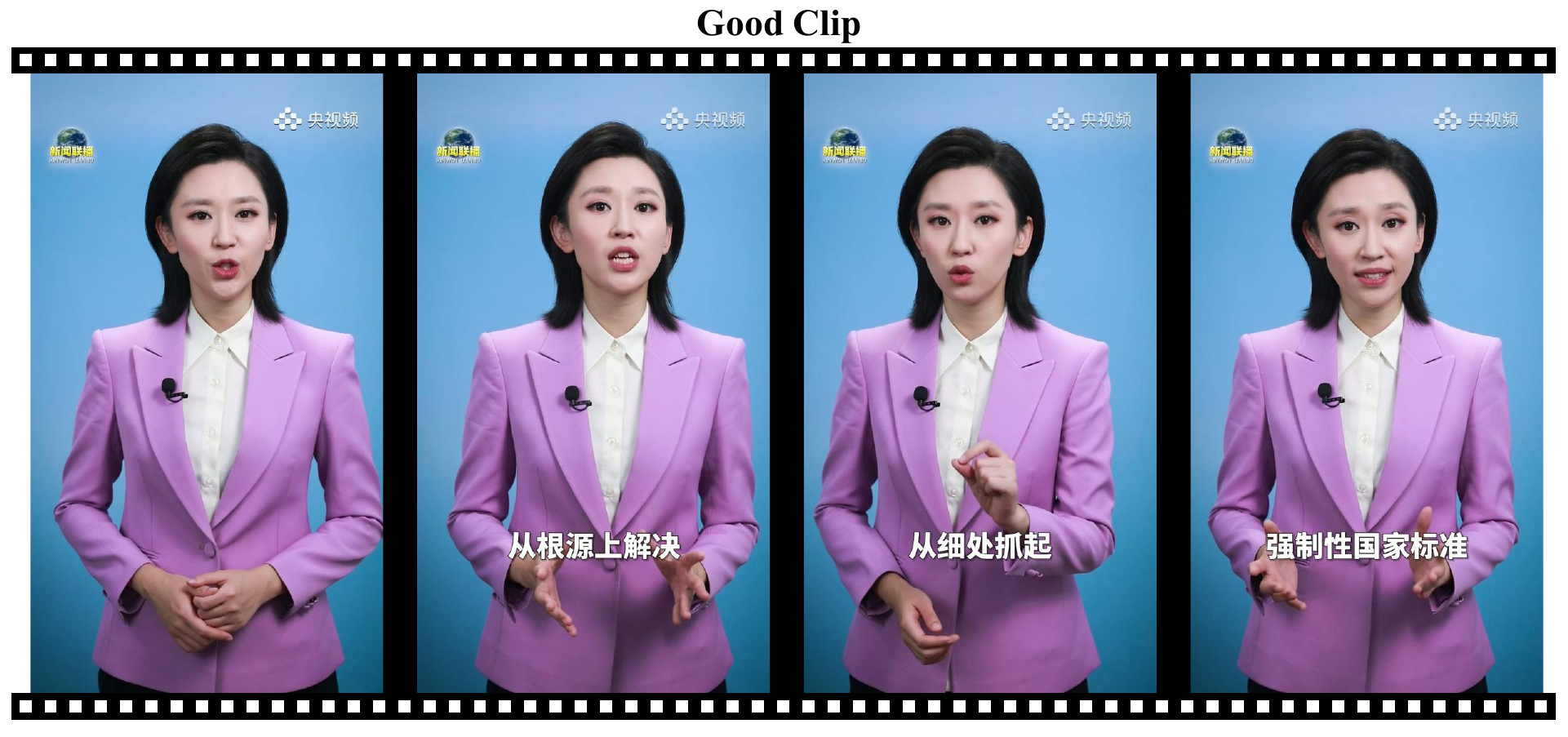}
        \caption{\textbf{Good Clip.} The video clip consistently features the same clearly visible person throughout, with rich facial expressions and body motions, except that the subtitles need to be removed in a subsequent processing step.}
       \label{fig:cas_good_clip}
    \end{figure}

    \begin{figure}[htbp]
      \centering
       \includegraphics[width=1\linewidth,page=2]{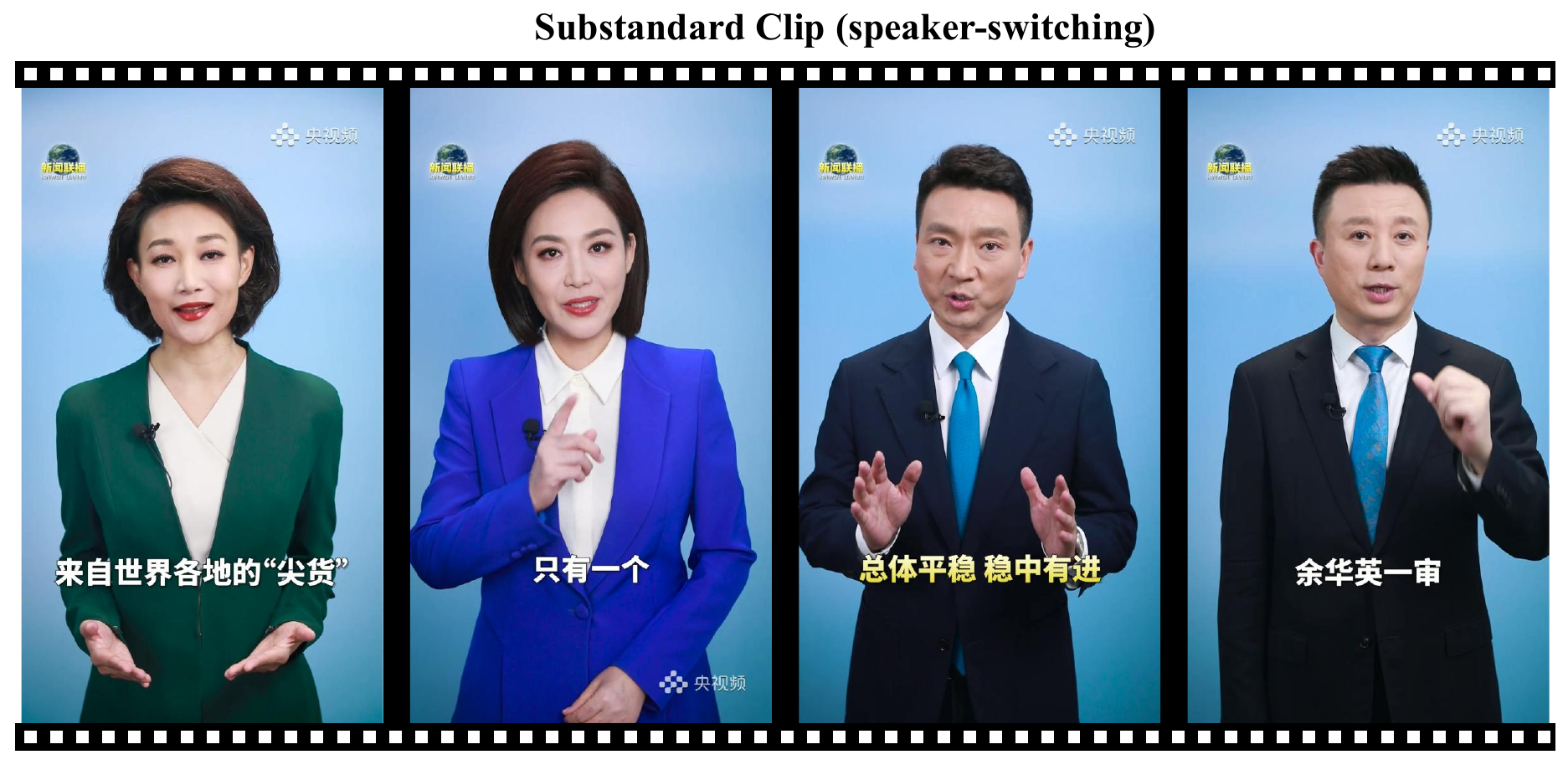}
      \includegraphics[width=1\linewidth,page=3]{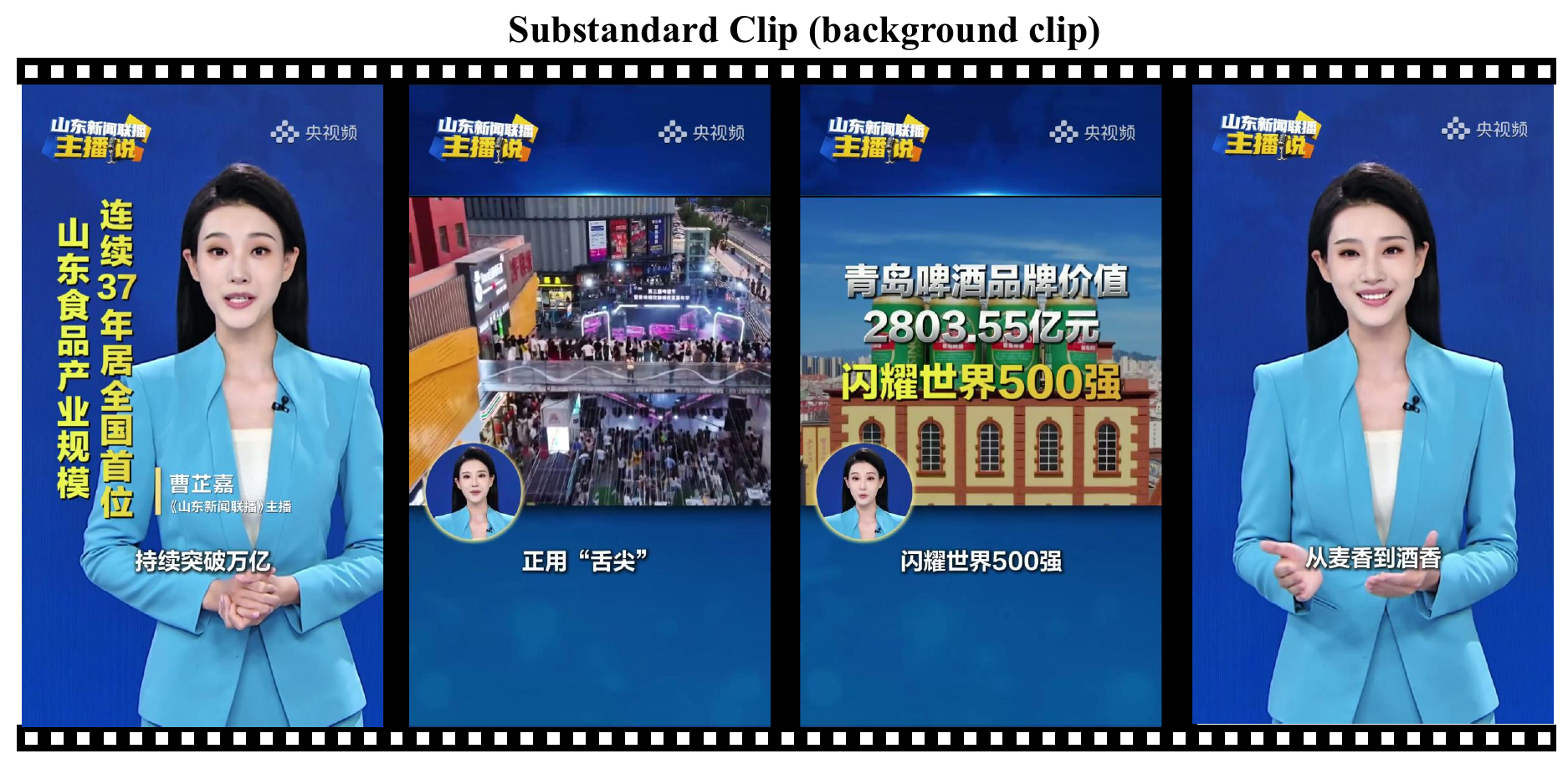}
      \includegraphics[width=1\linewidth,page=4]{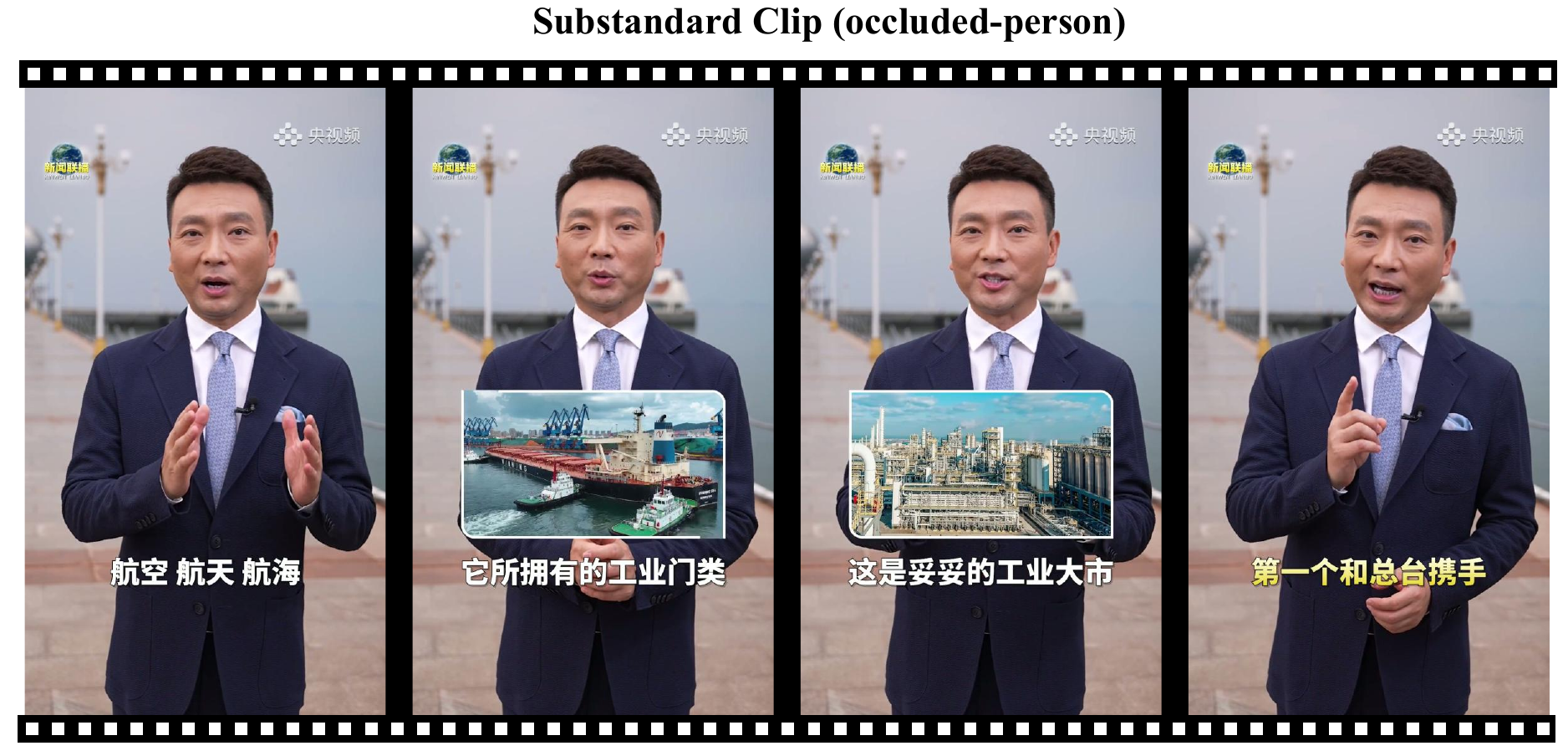}
       \caption{\textbf{Substandard Clip.} Further data processing is required to remove invalid frames and obtain video clips that meet the desired standards.}
       \label{fig:cas_bad_clip_1}
    \end{figure}

    \begin{figure}[htbp]
      \centering
       \includegraphics[width=1\linewidth,page=5]{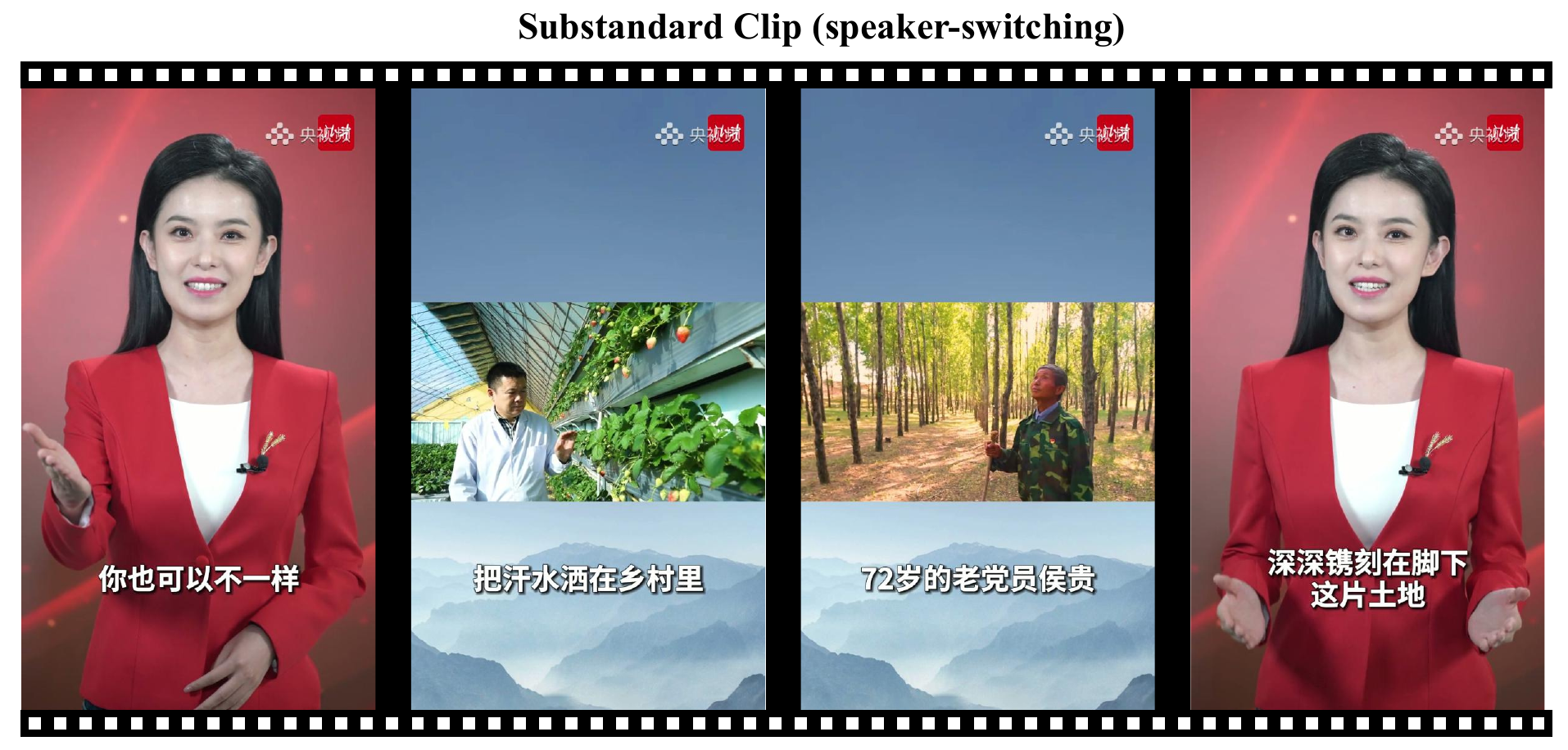}
      \includegraphics[width=1\linewidth,page=6]{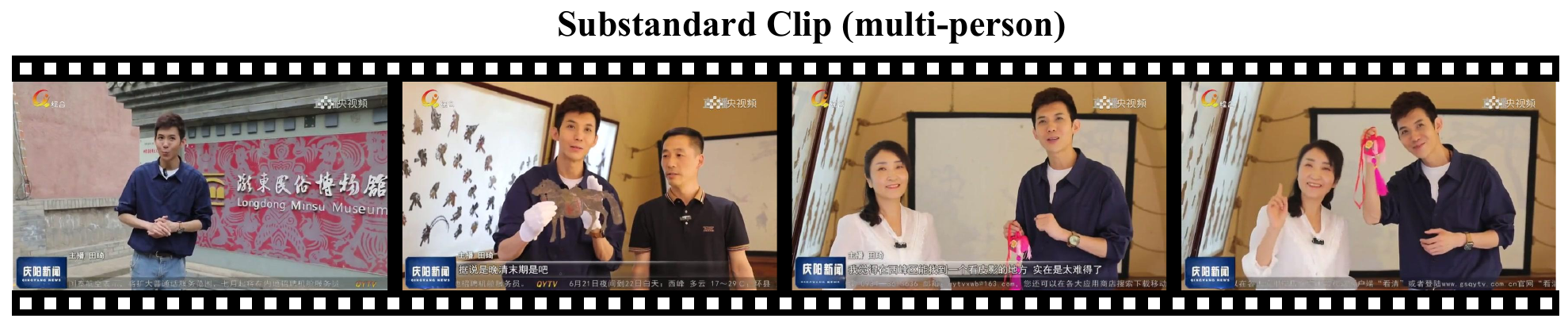}
      \includegraphics[width=1\linewidth,page=7]{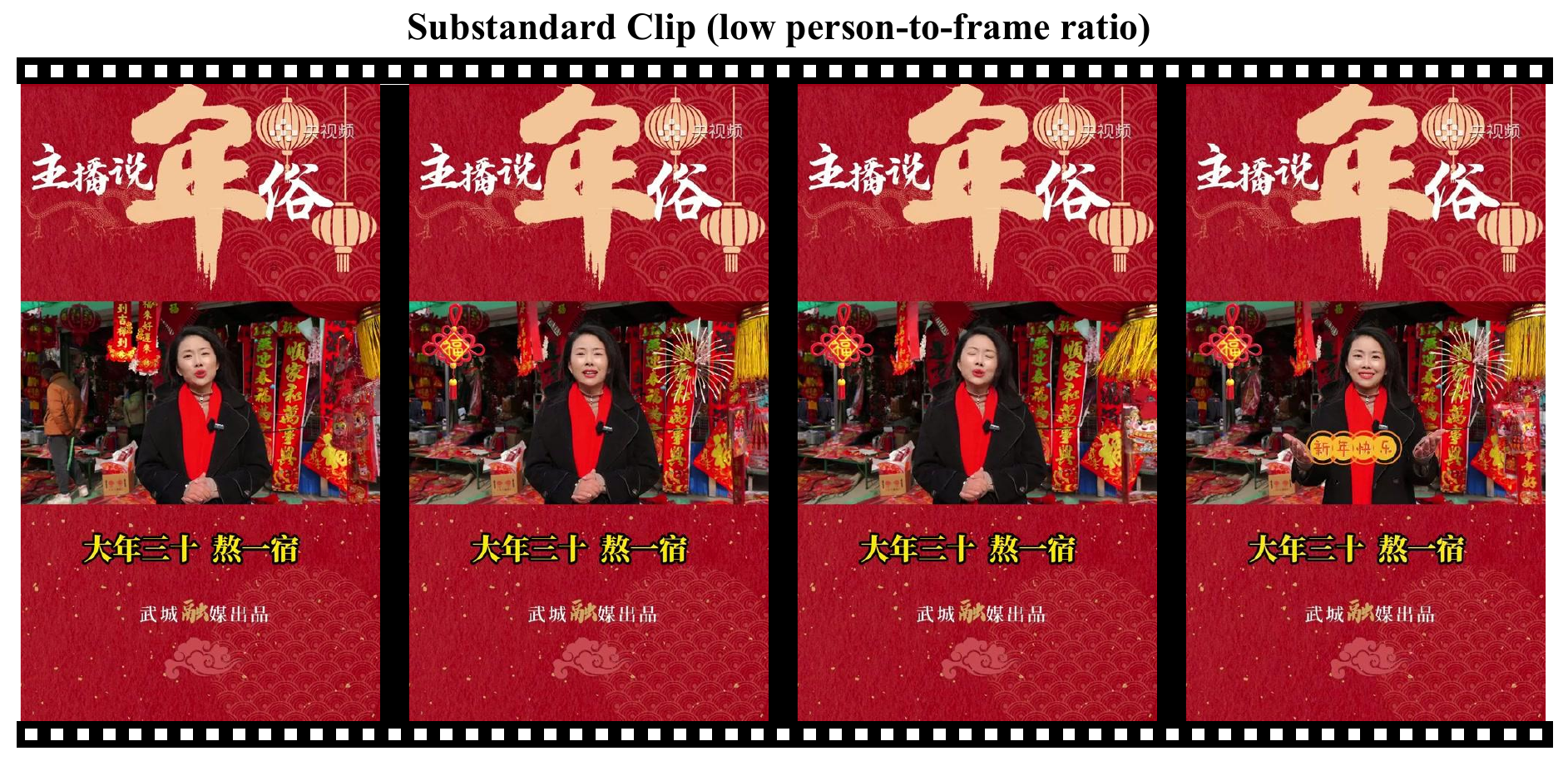}
      \includegraphics[width=1\linewidth,page=8]{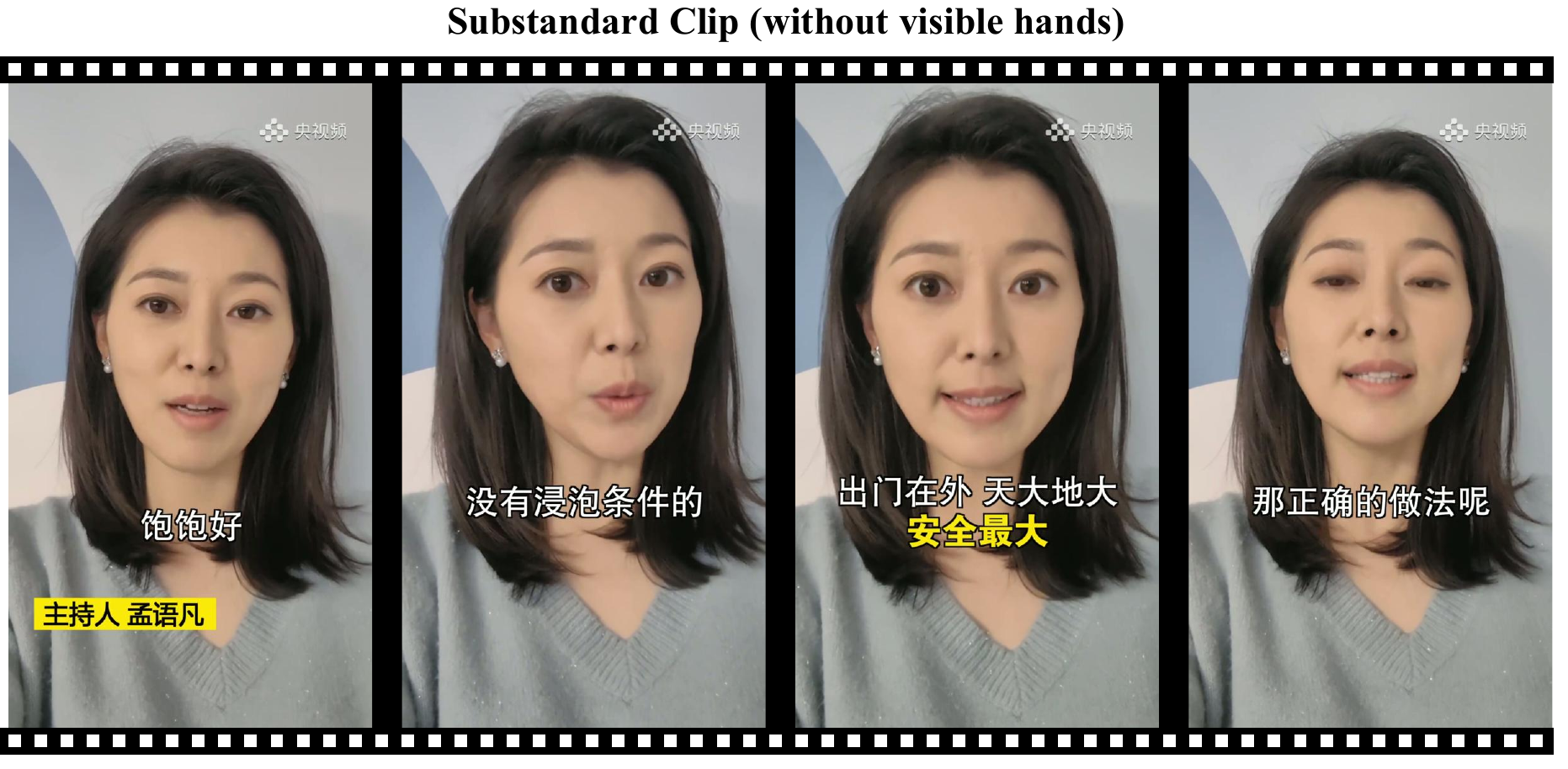}  
       \caption{\textbf{Substandard Clip.} Further data processing is required to remove invalid frames and obtain video clips that meet the desired standards.}
       \label{fig:cas_bad_clip_2}
    \end{figure}

    \clearpage
    
    \item \textbf{Frame sampling.} We sample one frame every 81 frames from each raw video to obtain candidate frames for subsequent filtering and identity reference selection.

    \item \textbf{Person filtering.} We first apply face detection~\cite{zhang2016mtcnn} to remove candidate frames without a visible face, thereby filtering out background-only frames and other frames without a clear speaking subject.
    
    \item \textbf{Identity reference image selection.} 

    For each original video, we select high-quality reference frames for each identity in three steps. First, we perform person detection~\cite{varghese2024yolov8} on candidate frames with detected faces and discard frames in which the detected person's bounding-box height is less than 70\% of the frame height. Second, we use Qwen2.5-VL\cite{bai2025qwen25vltechnicalreport} to identify obvious occlusions of the speaking subject and remove frames with noticeable occlusions. 
    Third, we extract face features with ArcFace\cite{deng2019arcface} and apply a DBSCAN-based clustering algorithm to group frames by identity. For each identity cluster, we retain the frame with the highest Qwen-VL-derived human-quality score as the identity reference image (\cref{fig:cas_filter_and_cluster}).
    Finally, all selected reference images are manually verified to ensure identity correctness and visual quality, and are then used as anchors for subsequent video temporal segmentation.

    \begin{figure*}[htbp]
      \centering
       \includegraphics[width=1\linewidth,page=1]{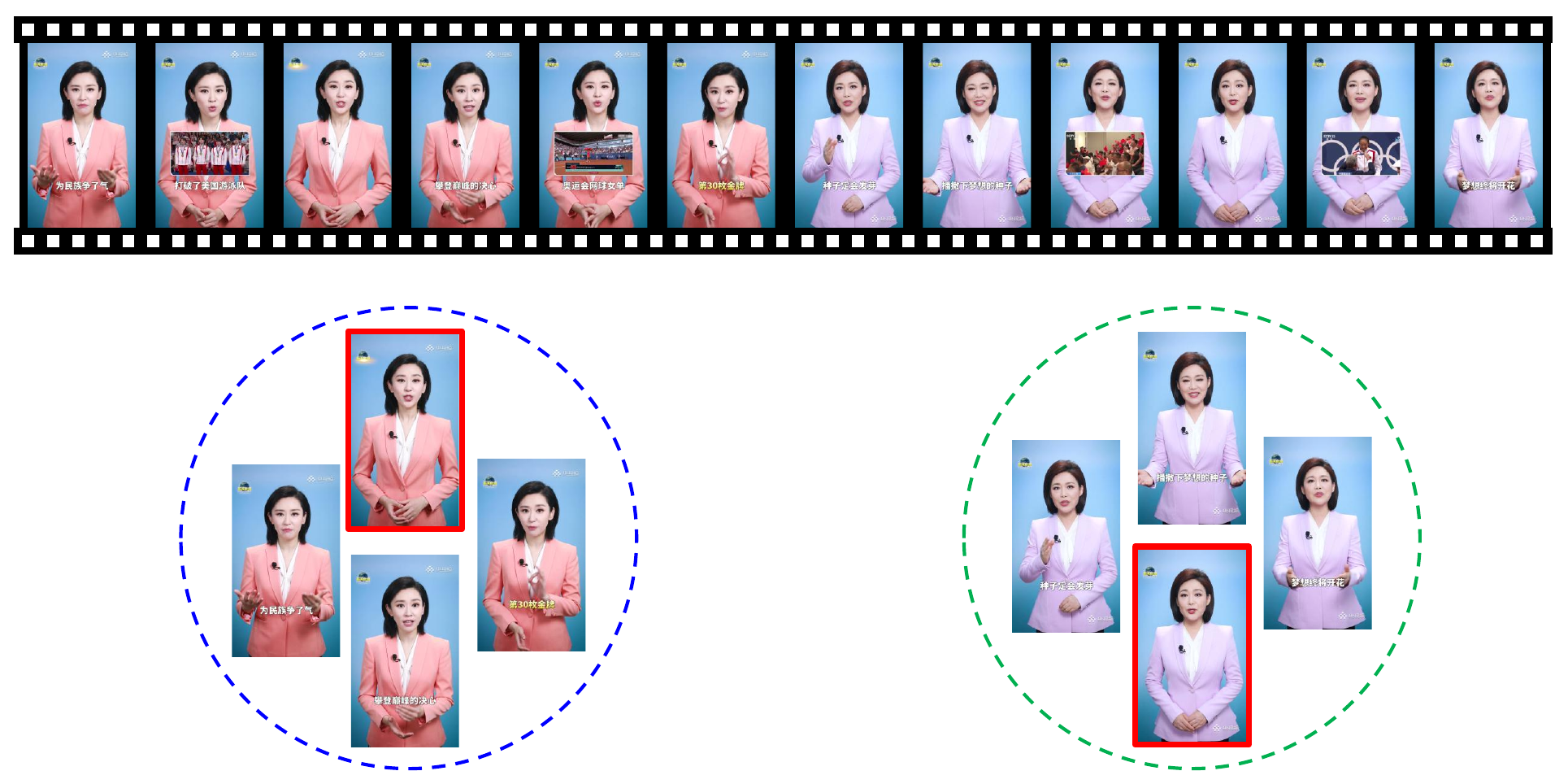}
        \caption{\textbf{Identity reference image selection.} After face clustering, the frames in each original video are grouped into multiple candidate identity clusters, each corresponding to a potential speaker identity. For each cluster, we select the frame with the highest Qwen-VL-derived human-quality score as the identity reference image, which serves as the anchor image (marked by a {\color[RGB]{255,0,0}red solid line} in the figure) for subsequent temporal segmentation.
    }
       \label{fig:cas_filter_and_cluster}
\end{figure*}

    \item \textbf{Video temporal segmentation.} 
    Since each raw video may contain multiple speakers, identity switches, or background clips, we segment it into continuous single-speaker chunks using person re-identification. Specifically, we use the selected identity reference images as anchors, extract person re-identification features with OSNet~\cite{zhou2021OSNet}, and assign each frame to the reference identity with the highest cosine similarity above a predefined threshold. Consecutive frames assigned to the same identity are grouped into a segment, and only segments longer than 81 frames are retained. The full algorithm is provided in Algorithm~\ref{alg:supp_CAS_Video_Seg}.

    \item \textbf{Subtitle removal.} 
    After temporal segmentation, we run OCR-based text detection and recognition~\cite{cui2025paddleocr} on all frames of each segmented clip to locate subtitle regions.
    Detected subtitle overlays are removed during preprocessing, while persistent static overlays, such as channel logos, are preserved because they do not affect gesture or identity consistency.
\end{enumerate}

\makeatletter\setlength{\@fptop}{7pt}\makeatother
\begin{algorithm}[t]
\caption{Video Temporal Segmentation via Identity Reference Image Similarity}
\label{alg:supp_CAS_Video_Seg}
\renewcommand{\baselinestretch}{1.15}\selectfont
\begin{algorithmic}[1]
\Statex \textbf{Variable definitions:}
\Statex $V = \{V_i \mid i = 0,\ldots, n-1\}$: set of source videos
\Statex $R_i = \{r_{i, j} \mid j = 0, \ldots, m_i-1 \}$: identity reference images for video $V_i$
\Statex $\mathrm{Similarity}$: cosine similarity of OSNet~\cite{zhou2021OSNet} features
\Statex $S$: similarity threshold
\Statex $W_i$: assigned reference index per frame in $V_i$
\Statex $L_\mathrm{min}$: minimum segment length (e.g., 81 frames)
\Statex $\text{Segments}$: continuous runs of frames with the same reference index and length $\geq L_\mathrm{min}$

\For{each video $V_i \in V$}
    \State $W_i \gets [\,]$
    \For{each frame $f_k$ in $V_i$, $k = 0$ to $N_i - 1$}
        \State $maxSim \gets 0$
        \State $selectedIdx \gets -1$
        \For{$r_{i,j} \in R_i$, $j = 0$ to $m_i - 1$}
            \State $sim \gets \mathrm{Similarity}(f_k, r_{i,j})$
            \If{$sim > S$ \textbf{and} $sim > maxSim$}
                \State $maxSim \gets sim$
                \State $selectedIdx \gets j$
            \EndIf
        \EndFor
        \State $W_i.\mathrm{append}(selectedIdx)$
    \EndFor

    \State $Segments \gets [\,]$
    \State $start \gets 0$
    \For{$k = 1$ to $|W_i|-1$}
        \If{$W_i[k] \neq W_i[start]$}
            \If{\resizebox{0.68\linewidth}{!}{$(k - start) \geq L_\mathrm{min}$ \textbf{and} $W_i[start] \neq -1$}}
            \State \resizebox{0.7\columnwidth}{!}{$Segments.\mathrm{append}([start,k-1,W_i[start]])$}
            \EndIf
            \State $start \gets k$
        \EndIf
    \EndFor
    \If{$(|W_i|-start) \geq L_\mathrm{min}$ \textbf{and} $W_i[start] \neq -1$}
        \State $Segments.\mathrm{append}([start, |W_i|-1, W_i[start]])$
    \EndIf

    \For{each $seg$ in $Segments$}
        \State \resizebox{0.9\linewidth}{!}{\textbf{Extract} a clip from $V_i$ using $seg[0]$ to $seg[1]$}
    \EndFor
\EndFor
\end{algorithmic}
\end{algorithm}

\onecolumn
\subsection{\textbf{Visualization.}}
\begin{figure*}[htbp]
    \centering
    \includegraphics[width=1\linewidth]{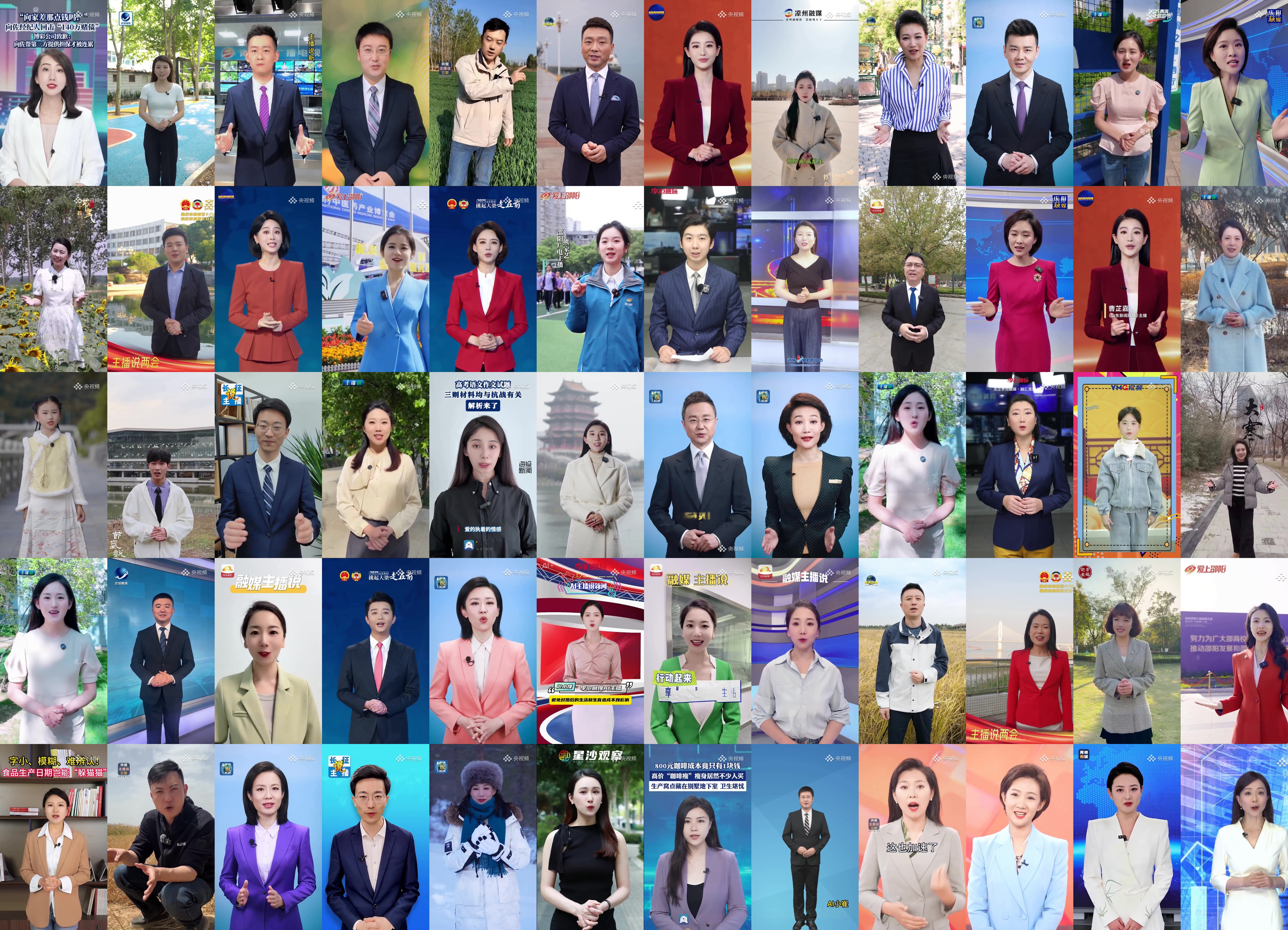}
    \caption{\textbf{Example frames from CAS.}
    CAS contains 19,052 curated clips spanning varied identities, poses, scene types, and camera viewpoints. This diversity supports robust evaluation of co-speech gesture generation, including facial articulation, upper-body dynamics, and global expressiveness.}
    \label{fig:CAS_show}
\end{figure*}

CAS includes diverse scenarios: indoor and outdoor settings, full- and half-body framing, standing and seated speakers, and both male and female anchors. Representative frames are shown in Fig.~\ref{fig:CAS_show}.

\twocolumn

\section{Limitations and Future Work}\label{sec:Limitations}

Although PAGE improves regional generation quality, local details, and audio--visual synchronization, several limitations remain. 
First, although PSR explicitly emphasizes important regions such as lips, eyes, hands, and finger gaps, the overall region partition is still relatively coarse. It does not fully model finer-grained structures and motions, such as individual finger articulation, subtle facial microexpressions, complex hand interactions or self-occlusions, expressive emotional changes, and challenging body poses such as side-facing or head-bowing motions. As a result, the model may still struggle to render highly complex gestures, expressions, and poses in a fully realistic manner.
Second, PAGE mainly assumes a fixed camera and presenter-style framing, which limits its ability to handle dynamic scenes, moving viewpoints, and interactive scenarios.
Third, under large distribution shifts, such as transferring from real portrait videos to cartoon-style portrait generation, the current negative samples may not fully cover all potential error patterns. In such cases, more diverse negative samples or additional calibration of the CCG classifier may be required.

In future work, we plan to extend PAGE toward more dynamic and expressive human animation, including full-body conversational scenarios where characters can talk while walking and interact with the surrounding environment.
We will also investigate multi-view consistency and improve the model's robustness to diverse viewpoints, poses, and human-object interactions.
In addition, we will explore more general and adaptive PSR weighting strategies, such as uncertainty-aware weighting, validation-feedback-based adjustment, and sample-adaptive region importance estimation, to improve robustness across diverse scenarios.
Finally, incorporating controllable speaking styles, emotions, and finer-grained motion representations will be an important direction for enhancing expressiveness, realism, and controllability, thereby broadening the applicability of PAGE to live media, virtual humans, and augmented reality.

%\balance
\bibliographystyle{IEEEtran}
\bibliography{IEEEfull}

\end{document}